\documentclass{article}
\usepackage[preprint]{neurips_2024}

\usepackage{hyperref}
\usepackage{microtype}
\usepackage{graphicx}
\usepackage{subcaption}
\usepackage{booktabs} % for professional table

\usepackage{amsmath}
\usepackage{amssymb}
\usepackage{mathtools}
\usepackage{amsthm}

\usepackage[capitalize,noabbrev]{cleveref}
\usepackage{verbatim}
\usepackage{subcaption}

\theoremstyle{plain}
\newtheorem{theorem}{Theorem}[section]
\newtheorem{proposition}[theorem]{Proposition}

\newtheorem{corollary}[theorem]{Corollary}
\theoremstyle{definition}

\theoremstyle{remark}

\newcommand{\R}{\mathbb{R}}

\newcommand{\E}{\mathbb{E}}

\usepackage[utf8]{inputenc} % allow utf-8 input
\usepackage[T1]{fontenc}    % use 8-bit T1 fonts
\usepackage{hyperref}       % hyperlinks
\usepackage{url}            % simple URL typesetting
\usepackage{booktabs}       % professional-quality tables
\usepackage{amsfonts}       % blackboard math symbols
\usepackage{nicefrac}       % compact symbols for 1/2, etc.
\usepackage{microtype}      % microtypography
\usepackage{xcolor}         % colors

\title{A Rational Model of Dimension-reduced Human Categorization}

\author{%
 Yifan Hong %\thanks{Use footnote for providing further information about author (webpage, alternative address)---\emph{not} for acknowledging funding agencies.} 
  \\
  Department of Industrial Engineering\\
  Tsinghua University\\
  Beijing, 100190 \\
  \texttt{hongyf23@mails.tsinghua.edu.cn} \\
  \AND
  Chen Wang \\
  Department of Industrial Engineering\\
  Tsinghua University\\
  Beijing, 100190 \\
  \texttt{chenwang@tsinghua.edu.cn} 
}

\begin{document}

\maketitle

\begin{abstract}
Humans can categorize with only a few samples despite the numerous features. 
To mimic this ability, we propose a novel dimension-reduced category representation using a mixture of probabilistic principal component analyzers (mPPCA).
Tests on the {\tt CIFAR-10H} dataset demonstrate that mPPCA with only a single principal component for each category effectively predicts human categorization of natural images. We further impose a hierarchical prior on mPPCA to account for new category generalization. mPPCA captures human behavior in our experiments on images with simple size-color combinations. 
We also provide sufficient and necessary conditions when reducing dimensions in categorization is rational.
\end{abstract}

\section{Introduction}\label{introduction}

Human categorization grasps commonalities across items despite their differences. 
Although natural stimuli have numerous features, people can {learn} new categories with just a few instances \citep{lake2015human} and {generalize} to novel observations \citep{salakhutdinov2012one,tiedemann2022one}. 
For example, a child can recognize a giraffe with only verbal descriptions. 
Theories suggest that people group instances with similar features together, and categories can be represented with past exemplars \citep{nosofsky1986attention} or abstract prototypes \citep{reed1972pattern}. Rational models \citep{anderson1991adaptive,griffiths2007unifying} provide a unifying perspective, casting categorization as optimal (Bayesian) inference.

These models are insightful, but they struggle to explain categorization in \textit{few-shot} settings where dimensions outnumber samples. 
For example, the rational model with full-rank covariance cannot obtain a reliable estimate directly, while exemplar-based approaches are naturally biased on unbalanced categories.
Moreover, dimensions with incidentally small variations can lead to poor model predictions on new samples \citep{pettine2023human}.

This paper proposes a novel dimension-reduced category representation under the rational framework. 
Each category is described by a prototype and a set of principal components (PCs), characterizing the location and within-category variations, respectively. 
On the natural image dataset with human labels {\tt CIFAR-10H} \citep{peterson2019human}, representation with merely a single principal component proves highly effective in predicting human categorization patterns while increasing dimensionality further leads to little improvement.

The dimension-reduced representation is compatible with a hierarchical prior over principal components. 
The resulting model, \textit{mixture of probabilistic principal component analyzers} (mPPCA) suggests a principled way of generalization in the few-shot setting. 
Within existing categories, principal components serve as low-dimensional local feature systems to locate subcategories. 
For a new category, mPPCA prefers to generalize along principal components of existing categories.
Behavioral experiments with simple visual patterns confirmed the anticipated generalization patterns, and mPPCA provides significantly better accuracy and correlation than classical models.

We also provide a theoretical rationale for dimension-reduced representation in human behavior. 
A dimension \textit{should} be preserved in the representation if and only if it provides relatively more information about within-category variation than category differences. 
Therefore, mPPCA mirrors human adaptation to the complex natural environment.

%Section 2 summarizes previous work. Section 3 introduces the PPCA representation for each category with theoretical analysis of when dimension-reduction improves categorization performance, followed by the hierarchical infinite mixture model in Section 4. We validate the model with simulations in Section 5 and behavioral experiments and tests on {\tt CIFAR-10H} in Section 6. Section 7 presents the discussion and limitation, and section 8 concludes the paper.

\section{Background}\label{sec:background}

\subsection{Models of human category learning}

Categorization groups instances with similar features. Category representations enable accurate predictions and consistent generalizations. Cognitive models of categorization make various assumptions about category representations. For example, the prototype model \citep{reed1972pattern} assumes that categories can be represented as abstract prototypes. People assign an instance to the category with probability proportional to the similarity to the prototype. The exemplar model \citep{nosofsky1986attention} considers a category to include all its known members. However, these classical models are confined to a fixed number of categories. 
The rational model of categorization (RMC) \citep{anderson1991adaptive} offers a different perspective. It postulates that human categorization results from adapting to the optimal prediction of features. RMC models categories as probability distributions and performs Bayesian density estimation. Denote $x_n\in\mathbb{R}^d$ and $c_n$ the new observation and its category assignment, respectively, and $\textbf{x}_{n-1}$ and $\textbf{c}_{n-1}$ the set of previous observations and their category memberships, respectively. The (posterior) predictive distribution of the features for a new observation is given by
\begin{equation}\label{eq:RMC_predictive}
\begin{split}
        P(x_n\vert \textbf{x}_{n-1}, \textbf{c}_{n-1})=\sum_{k=1}^K P(c_n=k\vert \textbf{c}_{n-1}) \cdot P(x_n\vert c_n=k, \textbf{x}_{n-1}, \textbf{c}_{n-1}).
\end{split}
\end{equation} 
The formulation decomposes the prediction task into two parts: the prior bias towards a particular category and the likelihood of an observation belonging to that category. The prior can take the form of a Chinese restaurant process (CRP) \citep{blackwell1973ferguson}, a sequential process that over $\textbf{c}_{n-1}$ that allows for infinite many categories. The probability of assigning a sample to an existing category $k$ is proportional to the number of existing samples $M_k$. Meanwhile, a new category emerges with probability proportional to a concentration parameter $\gamma > 0$:
\begin{equation}\label{eq-crp}
P(c_n=k\vert \textbf{c}_{n-1})\propto\left\{
\begin{aligned}
& M_k & & \text{if }  M_k > 0 ~(k \text{ is old}) \\
&\gamma & & \text{if }  M_k = 0 ~(k \text{ is new}) \\
\end{aligned}
\right.
\end{equation}
The CRP is the marginal distribution of category assignment corresponding to a Dirichlet Process (DP), which governs the joint distribution of category assignments and parameters for each category \citep{teh2010dirichlet}. DP has the constructive process known as the \textit{stick-breaking} construction \citep{blei2006variational}. For the prior probability measure $G$ of category parameter $\theta$ (without specifying the category), we have $G\sim DP(\gamma, H)$, and $G$ can be constructed as follows.  
\begin{equation}\label{eq:stick_breaking}
    \begin{split}
        \beta_k \sim \text{Beta}(1,\gamma), &\quad \theta_k\sim H,\\
        \pi_k = \beta_k\prod_{l=1}^{k-1}(1-\beta_l), &\quad G=\sum_{k=1}^\infty \pi_k\delta_{\theta_k}.
    \end{split}
\end{equation}
where $H$ is the base measure in the DP. The intuition is to sequentially sample for each category parameter $\theta_k$ a proportion $\beta_k$ from the remaining part of a stick (with $\sum_{k=1}^\infty \pi_k = 1$).

The likelihood in \cref{eq:RMC_predictive} can be a multivariate normal distribution for continuous variables with parameters $\theta_k = (\mu_k, \Sigma_k)$ for category $k$. The distribution specifies the mean parameters $\mu_k$ representing category prototypes and the covariance parameters $\Sigma_k$ defining dimensional variations.

RMC enjoys the flexibility to learn an indefinite number of categories. It can also be used in supervised and unsupervised settings and allows subcategory modeling \citep{griffiths2007unifying}. However, RMC models dimensional variations with full-rank covariance, and it is generally difficult to discern the similarity between covariances. Full-rank representation can also lead to degenerate performance with high-dimensional stimuli, as we will show later.

\subsection{Models of human generalization patterns}

Humans exhibit consistent generalization patterns in the feature space, e.g., isotropic or dimension-aligned \citep{smith1989model}. Through category learning, they gradually exhibit preferences to generalize along a meaningful axis, such as size or color. 
For example, \citet{shepard1987toward} uses $L_1$ metric and $L_2$ metric to describe generalization over different dimensions. 

In rational models, the covariance matrix reflects graded generalization that rotates and scales the feature space. It implies a direction of strong generalization through its first principal component. Researchers have imposed a mixture prior on the covariance matrix \citep{heller2009hierarchical} to highlight a preference to reuse dimensions for strong generalization. Consider a mixture of inverse Wishart distributions with $J$ components. Denote $\Phi_j$ as the parameters for the $j$-th component. The prior for the covariance matrix of category $k$ is given by $P(\Sigma_k\vert \Phi_1,\dots, \Phi_J) = \sum_{j=1}^J P(u_k=j)P(\Sigma_k\vert \Phi_j)$ where $u_k$ indicates which component to take effect. This model can also include infinitely many components using the CRP prior \citep{sanborn2021refresh}.

Notice that the covariance $\Sigma_k$ holds full-rank information about rotating and scaling the feature space. However, humans tend to focus only on a selective subset of dimensions when categorizing things \citep{aha1992concept}. Besides, the covariance implicitly determines the direction of strong generalization, making it challenging to identify subcategories. Therefore, a model that reduces the number of feature dimensions for each category can be favorable.

%Previous findings also imply a low-dimensional category representation, but in an implicit manner. 

\section{Dimension-reduced category representation}
\label{sec:PPCA}

To properly characterize human categorization, we need a combination of two elements: a hierarchical structure for generalization and a flexible local dimension-reduced representation. We start with dimension reduction for each category and then move on to a hierarchical model in the next section. %that allows categories to share feature dimensions for strong generalization. %Additionally, we discuss the implications of this model for identifying subcategories within each category.

%\subsection{Probabilistic PCA category representation} \label{model-PPCA}

We propose a low-dimensional representation of categories based on probabilistic principal component analysis (PPCA,  \citet{tipping1999probabilistic}). PPCA assumes that an observation $x_n\in \R^d$ is generated from a low-dimensional latent variable $z_n\in \R^q\,(q<d)$ with transformation
\begin{equation}
    x_n = W_{}z_n + \mu_{} + \epsilon_n.
\end{equation}
The columns of the \emph{loading} matrix $W\in \R^{d\times q}$ suggest the directions of strong generalization. The latent variable $z_n$ indicates variations in these directions. For convenience, we assume normal priors for the latent variables $z_n\sim N(0, I_q)$ and noises $\epsilon_n\sim N(0,\sigma^2I_d)$.

Denote $\{\theta_c\} = \{(\mu_c, W_c,\sigma_c^2)\}$ the parameters of all categories $c\in C$. Adopting a DP prior, after knowing $n-1$ observations $\textbf{x}_{n-1}$ and their category assignments $\textbf{c}_{n-1}$, we have the joint posterior
\begin{equation}\label{eq:CRP_posterior}
    \begin{split}
         P(\{\theta_c,\beta_c\} \vert \textbf{c}_{n-1}, \textbf{x}_{n-1}) \propto P(\{\theta_c\})P(\{\beta_c\})P(\textbf{c}_{n-1}\vert \{\beta_c\}) P(\textbf{x}_{n-1}\vert \textbf{c}_{n-1}, \{\theta_c\})
    \end{split}
\end{equation}
where $\{\beta_c\}$ come from the stick-breaking process, and $\{\theta_c\}$ are sampled from the base measure $H$ of the DP. The marginal posterior distribution of the category parameter $\theta$ is derived as
\begin{equation}\label{eq:CRP_posterior_theta}
    \begin{split}
         &P(\{\theta_c\}\vert \textbf{c}_{n-1}, \textbf{x}_{n-1})
         \propto P(\{\theta_c\})P(\textbf{x}_{n-1}\vert \textbf{c}_{n-1}, \{\theta_c\}),\\
         &P(\{\beta_c\} \vert \textbf{c}_{n-1}, \textbf{x}_{n-1})
         \propto P(\{\beta_c\})P(\textbf{c}_{n-1}\vert \{\beta_c\}),\\
         &\theta\vert \textbf{c}_{n-1}, \textbf{x}_{n-1} \sim G = \sum_{c \in C} \pi_c \delta_{\theta_c}
    \end{split}
\end{equation}
To formulate $H$, we assume independent normal for $\mu$ and multivariate normal for $W$. The prior for $W$ will be modified in \cref{sec:mPPCA} to incorporate shared principal component dimensions across categories.
The features of an observation $x_n$ given its category assignment $c_n$ follow the multivariate normal distribution $x_n\vert c_n, \theta_{c_n}\sim N(\mu_{c_n}, W_{c_n}W_{c_n}^T+\sigma_{c_n}^2I_d)$.
We then introduce the PPCA \emph{classifier} as the predictive distribution of category assignment $c_n$ given observation $x_n$
\begin{equation}\label{eq-CategoryAssign}
\begin{split}
    P(c_n\vert x_n, \textbf{c}_{n-1}, \textbf{x}_{n-1})=P(c_n\vert \textbf{c}_{n-1})P(x_n\vert c_n, \textbf{c}_{n-1}, \textbf{x}_{n-1}),
\end{split}
\end{equation}
where $P(c_n\vert \textbf{c}_{n-1})$ is easy to obtain based on the CRP process,  and the latter involves simulating the posterior of $\theta$,i.e. 
$
P(x_n\vert c_n, \textbf{c}_{n-1}, \textbf{x}_{n-1}) = \int_{\theta} P(x_n\vert c_n,\theta)dP(\theta\vert \textbf{c}_{n-1}, \textbf{x}_{n-1}).
$

Meanwhile, given category assignment $c_n$, the latent variable $z_n$ has the posterior
$z_n\vert x_n, c_n, \theta_{c_n} \sim N\big((W_{c_n}^TW_{c_n}+\sigma_{c_n}^2I_q)^{-1}W_{c_n}^T(x_n-\mu_{c_n}), \sigma_{c_n}^2(W_{c_n}^TW_{c_n}+\sigma_{c_n}^2I_q)^{-1}\big)$.
The principal components for each category span a low-dimensional feature system, with the latent variable $z_n$ explicitly capturing within-category variations. We explain how this relates to subcategory learning in  \cref{model-fs_generalization}. 

%\paragraph{Theoretical analysis of dimension reduction} When is a low-dimensional representation better than a full-rank one? Our result dimension-reduction helps when the dimension provides more information about between-category differences than within-category variations.

\subsection{Theoretical analysis of dimension reduction}
\label{sec:theory}

When is a low-dimensional representation better than a full-rank representation? We explore this question by considering the limiting case of PPCA when $\sigma^2\rightarrow 0$ so that it reduces to PCA and by focusing on two categories $C=\{a,b\}$. Observations from each category $c$ follow $x|c\sim N(\mu_c,\Sigma_c), \forall c\in C$. For simplicity, we assume equal covariance $\Sigma_a = \Sigma_b = \Sigma$, which is an important case in the real world. General covariance structure requires more complicated discussion, and may veil the clear intuition. The probability of assigning observation $x$ to the correct category (set to be $a$ without loss of generality) can be expressed by a sigmoid function 
\begin{equation}
    p(a\vert x) = \frac{e^{-\tau_q(x, \mu_a)}}{e^{-\tau_q(x, \mu_a)}+e^{-\tau_q(x, \mu_b)}}=\frac{1}{1+e^{-\{\tau_q(x,\mu_b)-\tau_q(x,\mu_a)\}}},
\end{equation} where $\tau_q$ is the projected distance to the subspace spanned by the first $q$ PC dimensions $(q<d)$, specified by eigenvectors $u_i, i = 1, \dots, q$ with decreasing eigenvalues $\lambda_1\geq...\geq \lambda_q$. The squared distance between category prototypes in the full-dimension space $r_{ab} = ||\mu_a-\mu_b||^2$ implies the amount of \textit{total information}, while $r_i= ||(\mu_a-\mu_b)^T u_i||^2$ describes the proportion of total information explained by the $i$-th PC (with $\sum_{i=1}^d r_i=r_{ab}$). We call $\alpha_q \triangleq \tau_q(x, \mu_b)-\tau_q(x, \mu_a)$ the \textit{sample discrimination index} for the $q$-dimensional PC subspace, reflecting how far the observation is from the wrong category relative to the correct one. 

We investigate when should the $(q+1)$-th PC dimension  be removed, and use the representation with the first $q$ PC dimensions. Define the signal-to-noise ratio (SNR) of the sample discrimination index
\[\text{SNR}_q=\frac{E_x[\alpha_q]^2}{Var_x[\alpha_q]},\quad q=0,1,...,d-2.\]
\cref{prop-PCA} presents the necessary and sufficient condition for excluding a PC dimension for the category representation to increase SNR. All proofs in this section are presented in the appendix.
\begin{proposition} \label{prop-PCA} 
 For given category prototypes $\mu_a,\mu_b$, discarding the $(q+1)$-th PC dimension (for $q=0,1,\dots,q-2$ from the category representation increases the signal-to-noise ratio of $\alpha_q$ ($\text{SNR}_{q}<\text{SNR}_{q+1}$) if and only if
    \begin{equation}\label{eq-SNR_condition}
    \lambda_{q+1} < \bigg(\frac{r_{q+1}}{\sum_{i=q+2}^d r_i}+2\bigg)\bigg(\frac{\sum_{i=q+2}^d r_i\lambda_i}{\sum_{i=q+2}^d r_i}\bigg).
    \end{equation}
\end{proposition}

The first term on the right-hand side reflects the information provided by the $(q+1)$-th PC dimension for differentiating categories. It suggests excluding a PC if it provides more information about cross-category variation than within-category variation. 
The intuition is confirmed with simulation in \cref{sec:representation_simulation} with PPCA classifier.
\cref{eq-SNR_condition} implies an improved performance bound in categorization, which is a monotone function of $\text{SNR}$. 

\begin{corollary}\label{cor-PCA}
     If \eqref{eq-SNR_condition} holds, dimension-reduction improves the accuracy lower bound for the PCA classifier.
\end{corollary}

\section{Hierarchical prior on feature dimensions}
\label{sec:mPPCA}
\subsection{Hierarchical infinite mixture of PPCA} \label{model-HIM_PPCA}

Now we present the mixture of PPCA (mPPCA), a nonparametric Bayesian hierarchical model based on the PPCA representation. It introduces dependencies between categories by sharing PCs. 
%\begin{figure}[ht]%
%\centering
%\includegraphics[width=0.6\linewidth]%{figs/HIM_PPCA_concept.pdf}
%\caption{Schematic illustration of the hierarchical prior over PCs. Categories share these components for common variation patterns.}\label{fig-HIM_PPCA_concept}
%\end{figure}

\begin{figure}[htbp]
\centering
    \begin{subfigure}{0.45\linewidth}
        \centering
        \includegraphics[width=\linewidth]{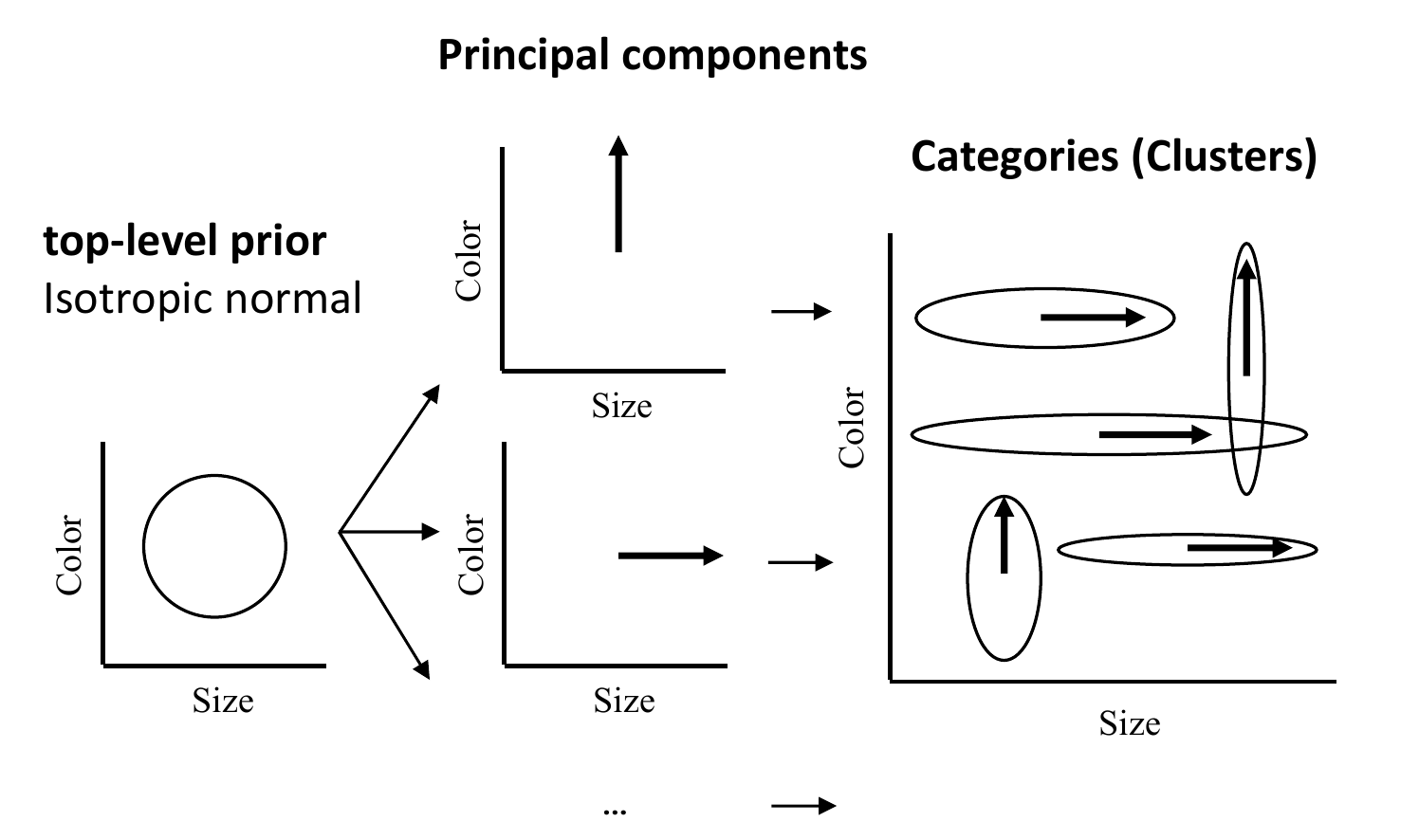}
        %\caption{new category}
        \label{fig-HIM_PPCA_concept}
    \end{subfigure}
    \begin{subfigure}{0.4\linewidth}
        \centering
        \includegraphics[width=\linewidth]{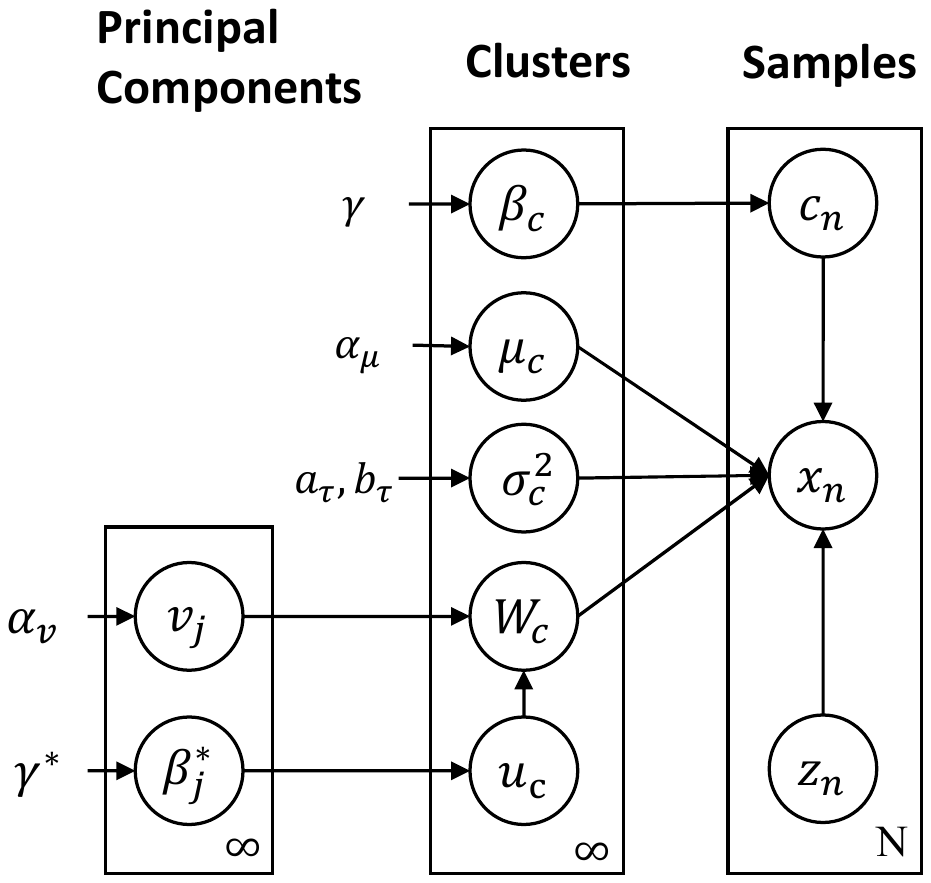}
        %\caption{subcategory}
        \label{fig:HIM_PPCA_model}
    \end{subfigure}
    \caption{(a) Schematic illustration of the hierarchical prior over PCs. Categories share these components for common variation patterns. (b) Graphical representation of mPPCA.}
    \label{fig1}
\end{figure}

There are two infinite mixtures in constructing the prior for mPPCA.
On the lower level, mPPCA describes observations as an infinite mixture of categories. The CRP prior \eqref{eq-crp} over category assignment allows infinitely many categories, but materializes only a finite set given the observations. Each category is represented by PPCA with its own parameters $\mu_c, W_c$. In this section, we assume each category has only one direction of strong generalization and is represented by a \emph{single} PC dimension.
On the higher level, we introduce another infinite mixture to share PC dimensions among categories. For each category $c$, an ownership indicator $u_c$ indexes a PC \textit{component} $\nu_{j}$ in the top-level mixture, modeled by the CRP prior. The chosen dimension $W_c$ is called the \textit{local} PC. Observations are drawn from the generative process presented in \cref{sec:generative_mPPCA}, with main idea presented in \cref{fig1}.

%\begin{figure}[ht]%
%\centering
%%\includegraphics[width=0.6\linewidth]{figs/HIM_PPCA_model.pdf}
%\caption{Graphical representation of mPPCA.}\label{fig-HIM_PPCA_model}
%\end{figure}

Adding a hierarchical prior changes the inference process. The posterior can be decomposed as the product of the conditional distribution of component-level parameters $\{\beta_j^*,\nu_j^*\}$ and the marginal of category-level parameters $\{\theta_c, \beta_c\}$.
\begin{equation}\label{eq:HIM_PPCA_posterior1}
\begin{split}
    p(\{\beta_j^*,\nu_j\}, \{\theta_c, \beta_c\} \vert \textbf{x}_{n-1}, \textbf{c}_{n-1}) = p({\beta_j^*,\nu_j}\vert \{\theta_c\}) p(\{\theta_c, \beta_c\}\vert \textbf{x}_{n-1}, \textbf{c}_{n-1})   
\end{split}
\end{equation}
where the second term is \cref{eq:CRP_posterior}. The first term is the posterior of the CRP mixture with concentration parameter $\gamma^*$ and normal base measure with covariance $\frac{1}{\alpha_\nu} I$,
\begin{equation}\label{eq:HIM_PPCA_posterior2}
    p(\{\beta_j^*, \nu_j\}\vert \{\theta_c\}) \propto p(\{\beta_j^*\}\vert \gamma^*) p(\nu_j\vert \alpha_\nu) \cdot p(\{u_c\}\vert \{\beta_j^*\})p(\{w_c\}\vert \{u_c\},\{\nu_j\}).
\end{equation}
\cref{eq:HIM_PPCA_posterior1} implies that the full posterior can be derived by the marginal posterior of category-level parameters and the conditional probability of component-level parameters.

Hierarchical prior in mPPCA supports consistent and flexible generalization. The component PCs of the top-level prior constitute an expressive set of feature dimensions shared among categories, resembling the central repository of features in humans \citep{austerweil2013nonparametric}. 

\subsection{Few-shot generalization}\label{model-fs_generalization}
mPPCA suggests a principled way of generalization in the few-shot setting. We consider generalization both within and beyond an existing category. 

\paragraph{Learning sub-categories} 
The principal components serve as a local feature system that supports subcategory learning. 
Consider a category $c$ with prototype $\mu_c$ and one local PC $w_c$. Let the latent variable be a probabilistic mixture of two components $z = sz_1 + (1-s)z_2$, where $s\sim \text{Bernoulli}(p_1)$ indicates which component is realized. Assuming standard normal prior for each component $z_1, z_2$, the marginal distribution of $z$ remains normal. 
A realization of latent variable $z_{sub}$ locates the subcategory prototype by utilizing the PC $w_c$ as a local feature system. The subcategory $x_{sub}\vert z_{sub} \sim N(\mu_c + w_{c} z_{sub}, \sigma^2 I)$ can be learned when no (full-dimensional) visual observation is available.

\paragraph{Learning new categories} Hierarchical prior guides generalizing of a new category. Learning over several observed categories leads to finite global PCs. Given only one sample $x_{new,1}$ from a new category, we cannot estimate a covariance directly. Hierarchical prior allows the new category to inherit generalization patterns from the existing ones. mPPCA suggests a category with mean $x_{new,1}$ and a PC $w_{new}$ sampled from the CRP posterior, with strong generalization along existing PCs. In a simple context where individuals learn to generalize along one certain direction $w_{new}$, a new category is represented as $x_{new}\vert w_{new}, x_{new,1} \sim N(x_{new,1}, \sigma^2I+w_{new}w_{new}^T)$. As a result, the new category can be learned with only one observation, which locates the new category prototype. 

\section{Simulation studies}

\subsection{Categorization with PPCA representation}\label{sec:representation_simulation}
We show that the optimal choice of dimensionality $q$ for category representation depends on the relative position of category prototypes and distribution of within-category variation on dimensions. 

\paragraph{Procedure} We generate 10000 samples from two categories $a,b$ in a 3-dimensional space. Category $a$ prototype is fixed at the origin, category $b$ prototype is specified with a unit vector and a scalar for the direction and distance. For covariance, we let the principal components align with the coordinates. The relative position of categories determines the information structure. The variance of the $1^{\text{st}}$ and $3^{\text{rd}}$ PC are fixed, while that of the second PC varies to control noise structure.

We consider three models, with $q=0,1,2$ for the category representation. $q=0$ implies the evaluation of Euclidean distance to category prototype. $q=1$ leads to a 1-dimensional PPCA representation. $q=2$ corresponds to a full-rank representation. For each model, $\sigma^2$ is set to the MLE, the average of variance of discarded dimensions in the representation $\sigma^2=\frac{1}{d-q}(\sum_{i=q+1}^d \lambda_i)$.

\begin{figure}[ht]
    \centering
    \includegraphics[width=0.45\linewidth]{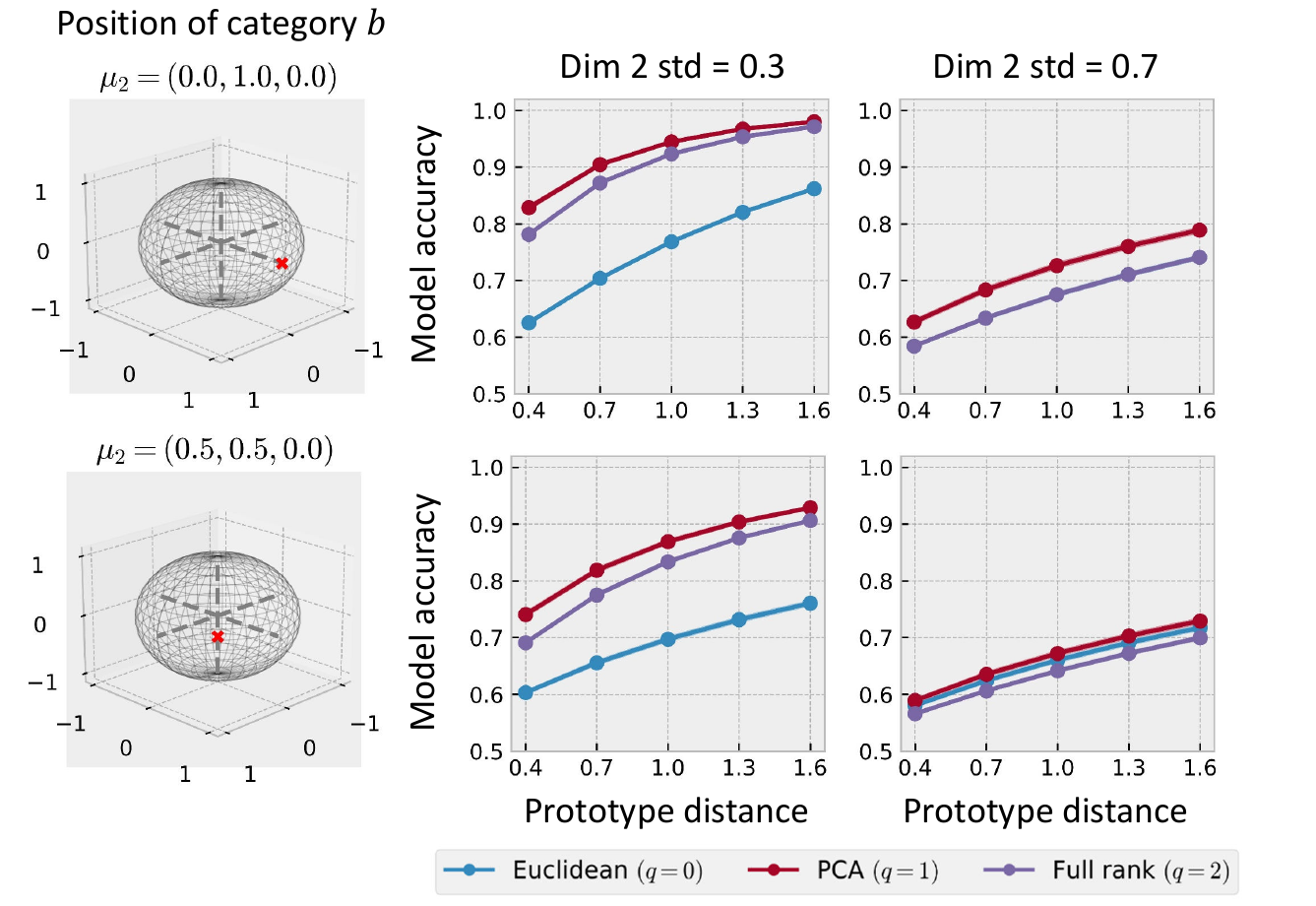}
    \caption{Model accuracy with varying prototype positions, distances and noise structure. When only dimension 2 is informative (the first row), rank-1 PPCA representation is optimal. For equally informative dimensions (the second row), rank-1 representation remains better. But when the noise levels become similar, the performance gap vanishes. (Full results are in \cref{sim-dimension_reduced_representation} in the Appendix.)} 
    \label{fig:sim1_main}
\end{figure}

\paragraph{Results} The results confirm the intuition given by \cref{sec:theory}. A dimension-reduced representation will improve accuracy when category prototypes differ on some removed dimensions. When only the first dimension is informative for categorization, removing it from the category representation is optimal (\cref{sim-dimension_reduced_representation} (a)). Results are similar when other dimensions are informative.
Meanwhile, when information is distributed uniformly on multiple dimensions, dimension reduction is effective when some dimensions do not reflect major variations within a category. %This finding is backed by the third setting (\cref{sim-dimension_reduced_representation} (c)), where all dimensions provide the same amount of information. 

\subsection{Hierarchical learning of generalization biases}\label{sec:mPPCA_simulation}

In this section, we illustrate the context-dependent learning of dimension-aligned generalization patterns. For mPPCA, PCs in the hierarchical mixture prior specify directions of strong generalization.

\paragraph{Procedure} The simulation involves two stages, a learning stage and a generalization stage. Training data is generated from a mixture distribution of axes-aligned categories, with large variance along one of the two dimensions. We also included a rotated version of categories. In the \textit{learning stage}, the model performs unsupervised learning on the training data. In the \textit{generalization stage}, a new stimulus is given, and we visualize the generalization pattern of the models by the probability of assigning any other new stimulus to the same category. We implement posterior inference using {\tt pyro} \citep{bingham2019pyro}. Details of model setup are shown in the Appendix.

\begin{figure*}[ht]%
\centering
\includegraphics[width=0.8\textwidth]{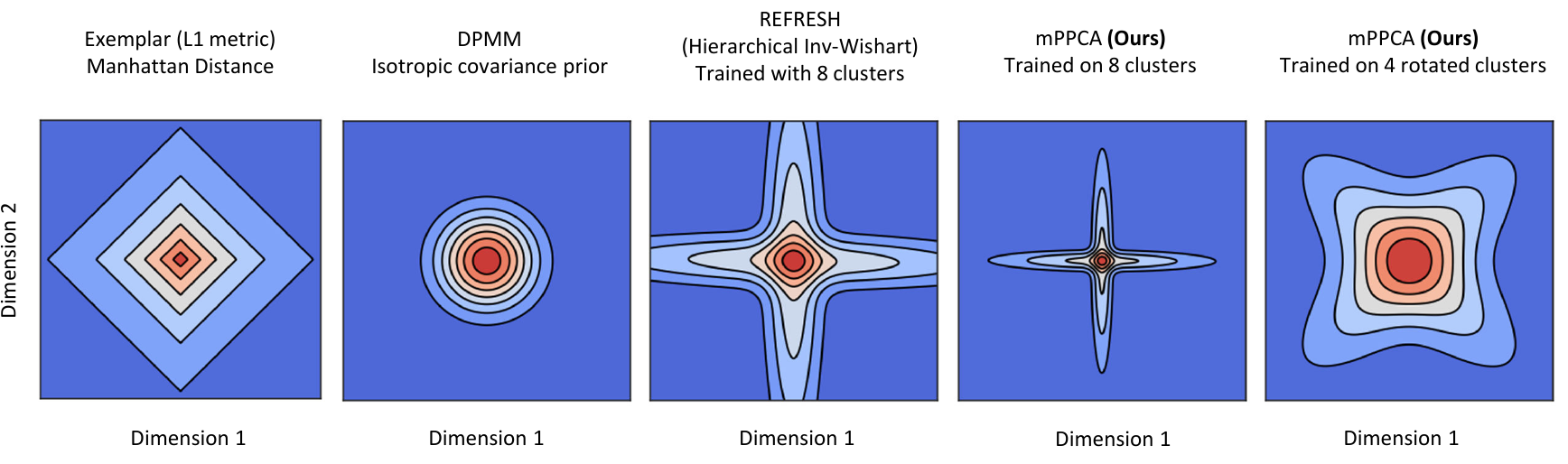}
\caption{Equal-generalization-probability contour of different models. The two axes correspond to dimensions in the psychological space (e.g., size and color).
After learning on axes-aligned clusters, hierarchical models (REFRESH and mPPCA) exhibit knowledge transfer.}\label{fig-Sim-generalization_grad}
\end{figure*}

\paragraph{Results} Since the training data contains balanced categories along each axis, we anticipate a mixture of strong generalizations along two dimensions. Model generalization patterns in \cref{fig-Sim-generalization_grad} show that mPPCA can learn the variations from the data. This resembles human generalization on separable dimensions \citep{sanborn2021refresh}, previously modeled as a mixture of covariance matrices (REFRESH). In comparison, the exemplar and DP mixture model cannot learn to generalize.

\section{Experiments}
\subsection{Categorization of natural images}\label{sec:DeepPPCA}

Low-dimensional stimuli are convenient for illustration but not realistic. 
To scale up, we explore human categorization of natural images using {\tt{CIFAR-10H}} \citep{peterson2019human}. 

\paragraph{Procedure} 
For each of the 10000 natural images in the test set, {\tt{CIFAR-10H}} includes 50 human categorization data. 
We use pre-trained convolutional networks as feature maps, including {\tt{ResNet18}} (512 dim), {\tt{Vgg11}} (512 dim) and {\tt{DenseNet121}} (1024 dim)\footnote{Models are adapted from https://github.com/huyvnphan/PyTorch\_CIFAR10 under the MIT licence.}. Their weights are held unchanged. 
We derive maximum likelihood estimate of classifier parameters on the training set (with hard labels), then compare the models on the test set. We compare mPPCA models with varying dimensions, and include the best performance of prototype and exemplar models with all three feature maps. 

\paragraph{Metrics} Besides accuracy, we record second best accuracy (SBA) and rank correlation with human data. SBA is the proportion of images on which the model predicts the second common human choice correctly. Rank correlation evaluates the ordinal associations between distributions. When models have similar accuracy, these metrics reflect the prediction of graded human categorization patterns. 

\begin{figure}[htb]
    \centering
    \includegraphics[width=\linewidth]{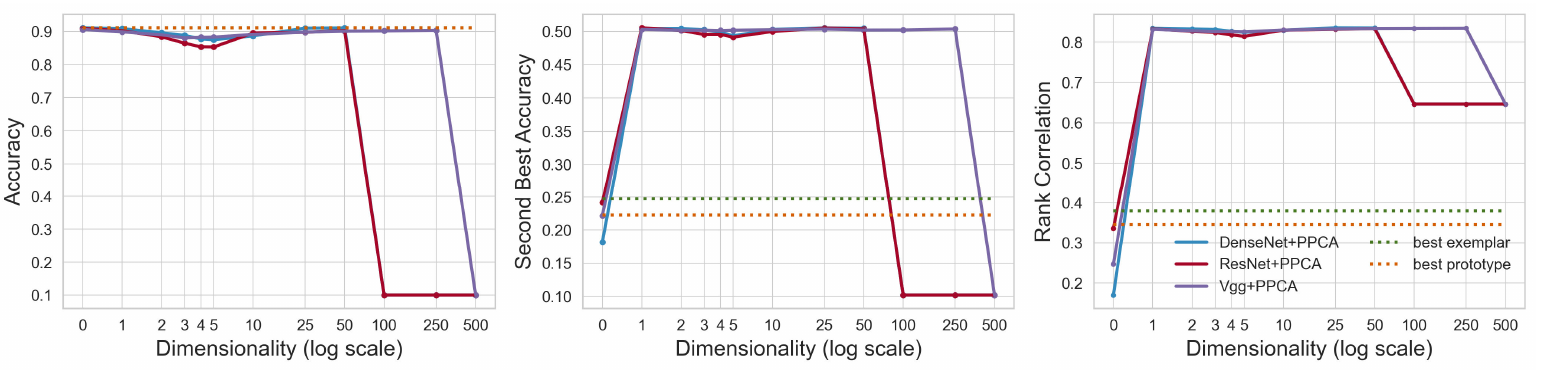}
    \caption{Model performance with different combinations of feature maps and classifiers. Since these measures are averaged over thousands of images, the error bars are negligible and are not included. }
    \label{fig:CIFAR_pre}
\end{figure}

\paragraph{Results} The results demonstrate that human categorization of natural images can be effectively captured with one principal component (\cref{fig:CIFAR_pre}). mPPCA model with a single PC in each category representation achieves impressive prediction performance on second best accuracy and rank correlation, surpassing both exemplar and prototype models. Meanwhile, increasing dimensionality does not further improve predictive power. Full-rank models even have degenerate performance. 

\subsection{Category few-shot generalization}\label{sec:exp-fs_generalization}

We carried out two experiments to study human few-shot generalization of a new subcategory or a new category. Artificial categories are used to avoid the confounding effect of human priors on learning, which may significantly influence the results. 

\begin{figure}[ht]%
\centering
\includegraphics[width=0.7\linewidth]{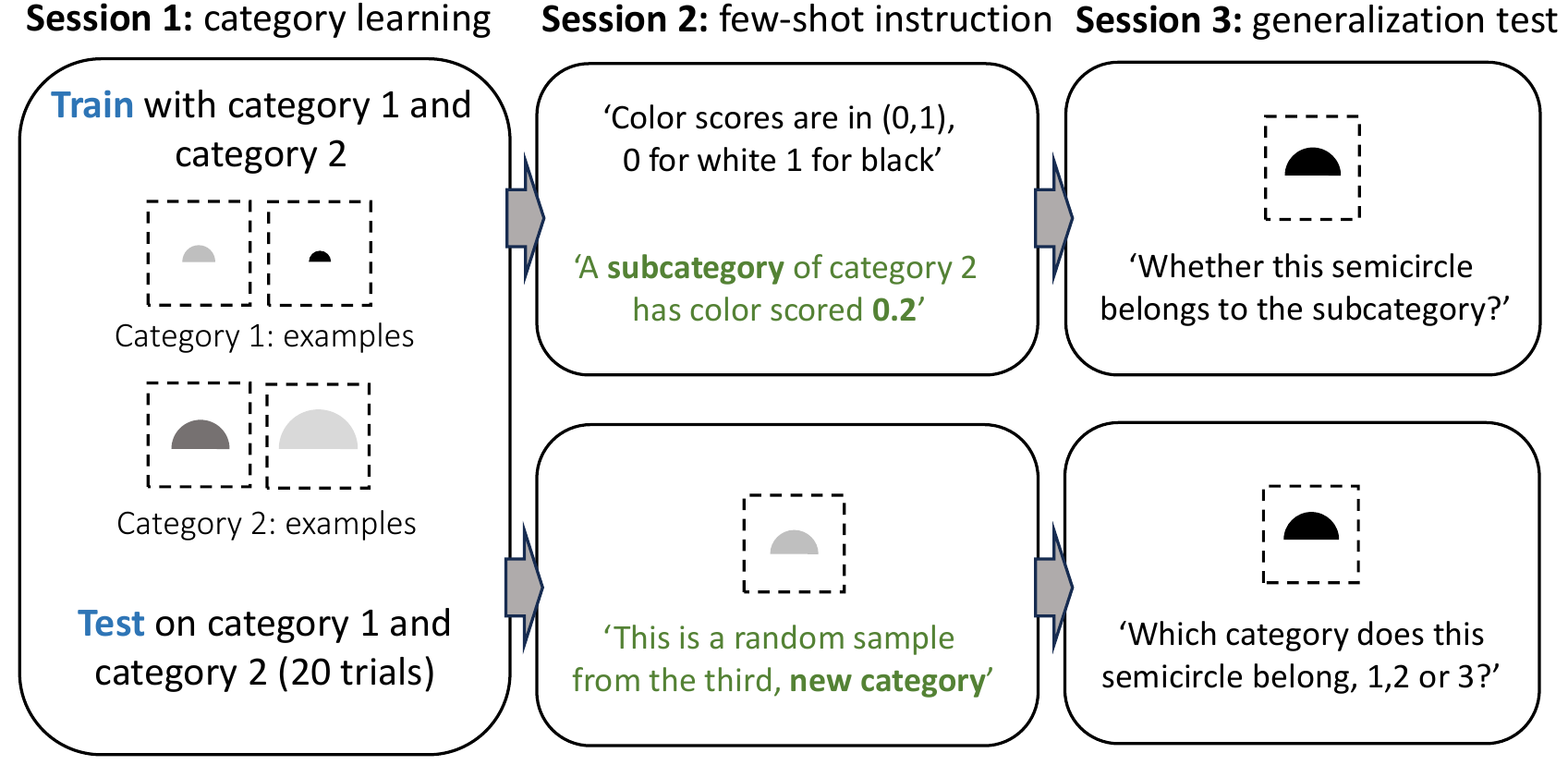}
\caption{Procedure of few-shot generalization experiment. Category 1 or 2 contains semicircles of regular size but varying color. After category learning in session 1, session 2 provides either one-shot or zero-shot instructions. The new category is similar to category 1 and 2 but locate near a different size value. The subcategory is generated from an isotropic Gaussian distribution, aligned with category 2 on the size dimension. Generalization patterns are tested in session 3.}\label{exp-fs_generalization-setting}
\end{figure}

\paragraph{Stimuli} The stimuli we choose are semicircles along varying \textit{color} and \textit{size} dimensions, two commonly used separable dimensions \citep{smith1989model, heller2009hierarchical}. We used dimension rating data to scale stimulus parameters based on perceived similarity. Each category corresponds to a multivariate normal distribution in the size-color space. The categories have small variances on the size dimension, and large variances on the color dimension. The subcategory has small variance on both dimensions. Stimuli are independent and identically sampled from the (sub)categories.

\paragraph{Procedure} Participants go through 3 sessions for each experiment: a category learning session containing the train and test phase, a few-shot instruction session, and a generalization test session. 
First, in the \textit{category learning session}, participants get familiar with the categories and their variations. They undergo training and testing phase, with 20 samples in each phase from category 1, 2 or neither. Training lasts until participants correctly categorize all the training samples. No feedback is available during the test. 
Second, in the \textit{few-shot instruction session}, participants learn about a new (sub)category. In the \textit{subcategory} experiment, verbal description of a subcategory is provided, describing its category PC (color) score. In the \textit{new category} experiment, one sample from the new category is provided. 
Third, in the \textit{generalization test session}, participants categorize 20 stimuli. Their choice is collected. 
The \textit{subcategory} experiment includes samples from one of the categories, some of which come from the subcategory. The participants judge whether the test stimuli come from the subcategory. In the \textit{new category} experiment, participants classify the samples into category 1, 2, or the new one. See \cref{exp-fs_generalization-setting} for an illustration.

We recruited 200 participants for each experiment on the online platform \textit{Credamo}, with 172 and 186 passing the attention tests, respectively. Participants undergo informed consent and are compensated fairly. 
Ethical issues are addressed carefully (See Impact Statement \cref{sec:impact_statement} for details). 
We compare mPPCA to prototype and exemplar models, with or without attention mechanism, and the rational model with necessary adaptations. Model setup is detailed in the Appendix.

\paragraph{Results} After training, most participants effectively learned the new (sub)category (\cref{exp-fs_generalization-session1-test}). Overall, in the few-shot setting, exemplar models fail to capture rapid learning of a new category. Prototype representations cannot generalize well with the attention mechanism. Rational model introduces clusters within categories, but suffers from identification of subcategory and covariance estimation for the new category. mPPCA provides a better account of human few-shot categorization.

%During \textit{generalization test}, mPPCA provides accurate predictions of human categorization. %Expected accuracy of prediction and correlation with participant choice are shown in \cref{exp-fs_generalization-sub_individual_metric} and \cref{exp-fs_generalization-new_individual_metric}.

\begin{figure}[ht]
\centering
    \begin{subfigure}{0.3\linewidth}
        \centering
        \includegraphics[width=\linewidth]{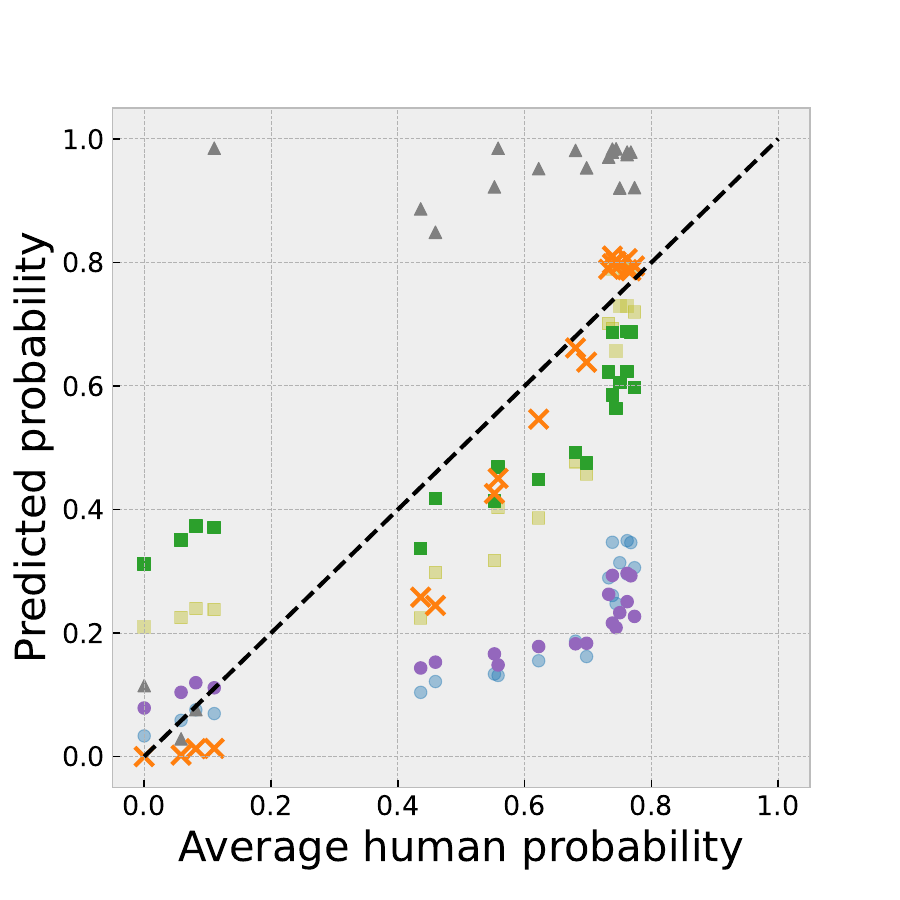}
        \caption{subcategory}\label{exp-fs_generalization-ProbPred_sub}
    \end{subfigure}
    \begin{subfigure}{0.3\linewidth}
        \centering
        \includegraphics[width=\linewidth]{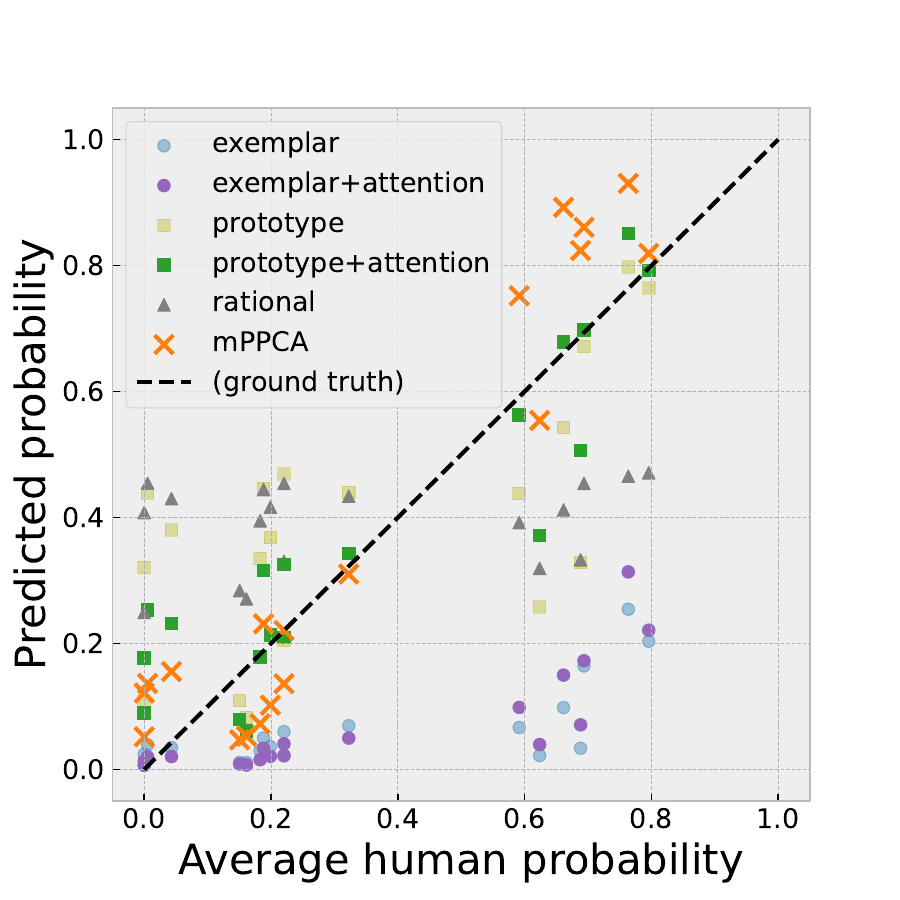}
        \caption{new category}\label{exp-fs_generalization-ProbPred_new}
    \end{subfigure}
    \caption{Model prediction of human assignment probability to the new (sub)category.}
    \label{exp-fs_generalization-ProbPred}
\end{figure}

In the subcategory experiment, mPPCA produces predictions with significantly higher accuracy and correlation with human choice (\cref{table-fs_generalization}). Its generalization pattern matches human behavior (\cref{exp-fs_generalization-sub_pred}). Both exemplar models and prototype models, with or without attention mechanism, underestimate the probability of the subcategory (\cref{exp-fs_generalization-ProbPred_sub}). Rational model has difficulty identifying the subcategory, since the clusters in the category are not necessarily identical with the new subcategory. It's worth mentioning that the attention mechanism hurts categorization performance within a category. Instead, humans adopt flexible context-dependent feature weighting, which is captured by mPPCA.

In the new category experiment, exemplar models, affected by unbalanced categories, underestimate the probability of the new category (\cref{exp-fs_generalization-ProbPred_new}). Prototype models (with or without attention) produce similar generalization patterns as mPPCA (\cref{exp-fs_generalization-heatmap}), both providing good predictions. Rational model cannot estimate the covariance of the new category, as only cluster-level parameters are available. Using Euclidean distances for categorization, its generalization pattern deviates from human behavior (\cref{exp-fs_generalization-new_pred}).
mPPCA predicts human category assignments more accurately (\cref{table-fs_generalization}).

\begin{table*}[ht]
  \caption{Performance in predicting human few-shot generalization of subcategories and new categories}
  \label{table-fs_generalization}
  \centering
  \begin{tabular}{lcccc}
    \toprule
    &\multicolumn{2}{c}{subcategory learning} & \multicolumn{2}{c}{new category learning}     \\
    \cmidrule(r){2-3}
    \cmidrule(r){4-5}
    Model  & expected accuracy & correlation & expected accuracy & correlation \\
    \midrule
    Exemplar & 0.517$\pm$0.043 & -0.102$\pm$0.118 &  0.594$\pm$0.063  & 0.372$\pm$0.128 \\
    Exemplar+Attention & 0.498$\pm$0.041 & -0.102$\pm$0.118 & 0.620$\pm$0.064 & 0.407$\pm$0.117\\
    Prototype & 0.599$\pm$0.020 &  0.351$\pm$0.091 & 0.562$\pm$0.030 & 0.607$\pm$0.051 \\
    Prototype + Attention & 0.555$\pm$0.012 &  0.351$\pm$0.091 &0.638$\pm$0.030 & \textbf{0.688$\pm$0.044} \\
    Rational model & \textbf{0.668$\pm$0.039} &  0.374$\pm$0.068 & 0.467$\pm$0.019 & 0.570$\pm$0.044 \\
    mPPCA (\textbf{Ours}) & \textbf{0.662$\pm$0.033} & 
 \textbf{0.451$\pm$0.065} & \textbf{0.705$\pm$0.028}  & \textbf{0.696$\pm$0.040} \\
    \bottomrule
  \end{tabular}
  \end{table*}

\section{Discussion and limitation}\label{sec:discussion} %% ++
Properties of PPCA makes it possible to explain cross-categorization and context-dependent behavior. PPCA does not impose orderings among PCs, enabling context-dependent ordering of features, as in cross-categorization \citep{shafto2011probabilistic}. Besides, PPCA does not assume orthogonality and can learn correlated features, similar to human feature learning. For example, saturation and brightness, two correlated color dimensions, are learned by color experts \citep{austerweil2010learning}. mPPCA implies a two-level structure of categories and subcategories. Its relation with structural organization of the categories \citep{canini2011nonparametric} is a promising future work.

Humans can also learn categories in other ways, like social learning. It's unrealistic that all principal components in human minds are learned through direct observation. This assumption is made for simplification purposes. Other aspects of human learning are out of the scope of this paper. 
\section{Conclusion}

We propose mPPCA, a flexible generalization of previous rational models of categorization with dimension-reduced category representation. Such a low-dimensional representation benefits categorization in a noisy environment, where certain dimensions provide more information about category differences than about internal variations. Simulations verify our theoretical findings and illustrate the model behavior. mPPCA model reproduces human-like categorization on {\tt CIFAR-10H} natural images, and can effectively capture human few-shot generalization within or beyond categories. 

\begin{ack}
This work was supported by the National Natural Science Foundation of China (NSFC) 72192824.
\end{ack}

%\section*{References}

\bibliographystyle{apalike}
{
\small
\bibliography{ref}
}

%%%%%%%%%%%%%%%%%%%%%%%%%%%%%%%%%%%%%%%%%%%%%%%%%%%%%%%%%%%%

\appendix

\newpage
\section{Appendix: model and theory}

\subsection{Generative process of mPPCA}\label{sec:generative_mPPCA}

\begin{itemize}
    \item [(1)] For each component in the higher-level mixture,
        \begin{itemize}
            \item [(a)] Draw probabilistic PC $\nu_j\sim N(0,\frac{1}{\alpha_\nu}I_d)$.
            \item [(b)] Draw stick-breaking weight $\beta_j^*\sim \text{Beta}(1,\gamma^*),\, \pi_j^* = \beta_j^*\prod_{i=1}^{j-1}(1-\beta_j)$.
        \end{itemize}
    \item[(2)] For each category in the lower-level mixture,
    \begin{itemize}
        \item [(a)] Draw component assignment $u_c \sim \text{Mult}(\{\pi_j^*\})$. $w_c = \nu_{u_c}+\xi_c$, where $\xi_c$ is a normal noise term.
        \item[(b)] Draw category prototype $\mu_c\sim N(0,\frac{1}{\alpha_\mu}I_d)$.
        \item [(c)] Draw stick-breaking weight $\beta_c\sim \text{Beta}(1,\gamma),\, \pi_c = \beta_c\prod_{l=1}^{c-1}(1-\beta_c)$.
        \item[(d)] Draw noise variance $\sigma_c^2\sim \text{Inv-Gamma}(a_\tau, b_\tau)$.
    \end{itemize}
    \item [(3)] For each sample $x_n,\, n=1,...,N$, 
    \begin{itemize}
        \item [(a)] Draw category assignment $c_n\sim \text{Mult}(\{\pi_c\})$
        \item [(b)] Draw latent variable $z_n\sim N(0,1)$.
        \item [(c)] Draw observation
        \[x_n\vert z_n, c_n \sim N(\mu_{c_n} + w_{c_n}z_n, \sigma_c^2 I_d)\]
    \end{itemize}
\end{itemize}

\subsection{Proofs for the theoretical analysis}\label{sec:proof}
\paragraph{Proof for \cref{prop-PCA}}
\begin{proof}
    We first formalize some useful notations. The covariance matrix of the categories has eigen-decomposition $\Sigma_c=U\Lambda U^T$, where the diagonal matrix $\Lambda = \text{diag}(\lambda_1,...,\lambda_d)$ consists of the eigenvalues, and columns of $U$ are corresponding eigenvectors. The truncated matrix containing first $q$ columns of $U$ is denoted as $U_q$, with corresponding $\Lambda_q\text{diag}(\lambda_1,...,\lambda_q),\, q<d$. Linear projection into the subspace is represented  as $P=W(W^TW)^{-1}W^T=U_qU_q^T$. 
    
    According to the definition, $\alpha \triangleq ||(I-P)(x-\mu_b)||^2-||(I-P)(x-\mu_a)||^2$. 
    For any given category prototypes, $\mu_a,\mu_b$, and projection matrix $P=U_qU_q^T$, the expectation and variance of the sample distinction index $\alpha$ can be derived as
    \begin{equation}
        \E_x[\alpha] = ||(I-P)(\mu_a-\mu_b)||^2 = r_{ab}-\sum_{i=1}^q r_i,
    \end{equation}
    \begin{equation}
        \text{Var}_x[\alpha] = 4(\mu_a-\mu_b)^T\Sigma_c(\mu_a-\mu_b) - 4((\mu_a-\mu_b)^TU_q\Lambda_q U_q^T(\mu_a-\mu_b))=4\sum_{i=q+1}^d \lambda_i r_i.
    \end{equation}
    When considering distance to the principal subspaces spanned by the first $q$ eigenvectors, the signal-to-noise ratio of $\alpha$
    \begin{equation}\label{eq-SNR}
        \text{SNR}_q =\frac{\E_x[\alpha]^2}{var_x(\alpha)} = \frac{1}{4}\frac{(r_{ab}-\sum_{i=1}^q r_i)^2}{\sum_{i=q+1}^d \lambda_i r_i}
    \end{equation}
    Hence, the decision to exclude dimension $q+1$ will increase signal-to-noise ratio ($\text{SNR}_{q+1}>\text{SNR}_q$) if and only if $\lambda_{q+1}<\frac{2r_{ab}-2\sum_{i=1}^{q+1} r_i + r_{q+1}}{(r_{ab}-\sum_{i=1}^{q+1} r_i)^2} \sum_{i=q+2}^d r_i\lambda_i$, which leads to inequality \ref{eq-SNR_condition} with minor transformation.
\end{proof}

\paragraph{Proof for \cref{cor-PCA}}
\begin{proof}
    PCA corresponds to the limit of PPCA as $\sigma^2\rightarrow 0$. Hence, the classifier chooses  with probability 1 the category whose principal subspace is the closest. This leads to $p(\hat{y}=a\vert a,b)=p(\alpha >0\vert a,b)$.
    From one-sided Chebyshev's inequality,
    \begin{equation}
        P(\alpha >0\vert a,b)\geq \frac{E_x[\alpha]^2}{Var(\alpha\vert a,b)+E_{x}[\alpha]^2} = \frac{\text{SNR}}{1+\text{SNR}}
    \end{equation}
    Since it is a monotonic function of signal-to-noise ratio, we immediately arrives at the corollary.
\end{proof}

\section{Appendix: simulation and experiment details}
\subsection{Details of simulation study: categorization with PPCA representation}

\begin{figure}[htbp]
    \centering
    \begin{subfigure}{0.9\textwidth}
    \centering
        \includegraphics[width=\linewidth]{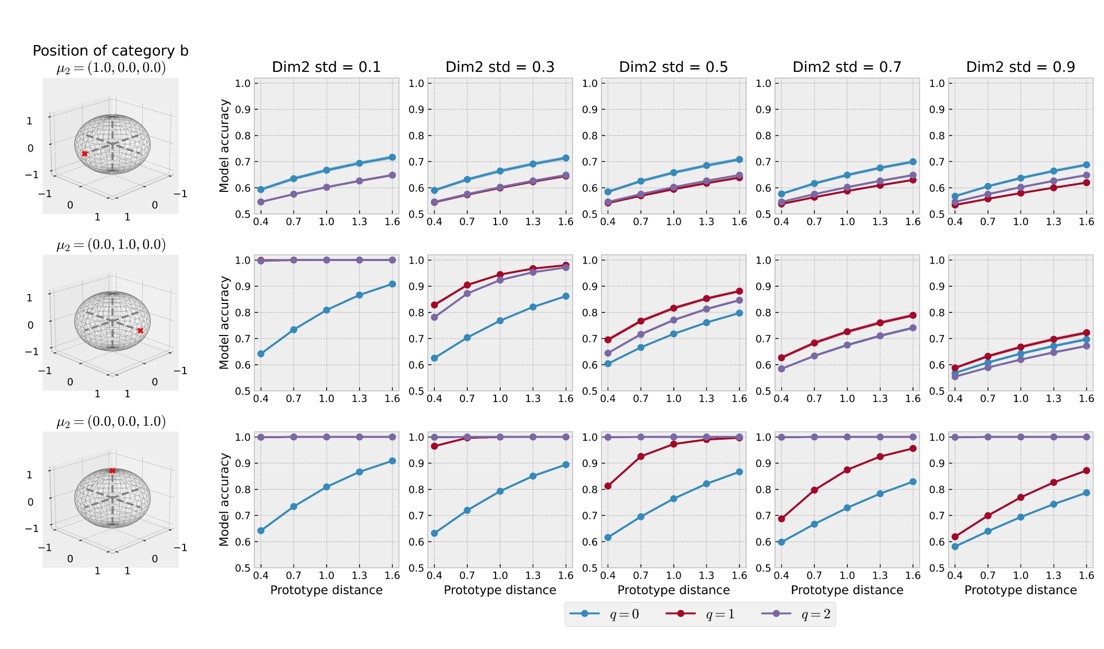}
        \caption{One informative dimension}
    \end{subfigure}
    
    \begin{subfigure}{0.9\textwidth}
        \centering
        \includegraphics[width=\linewidth]{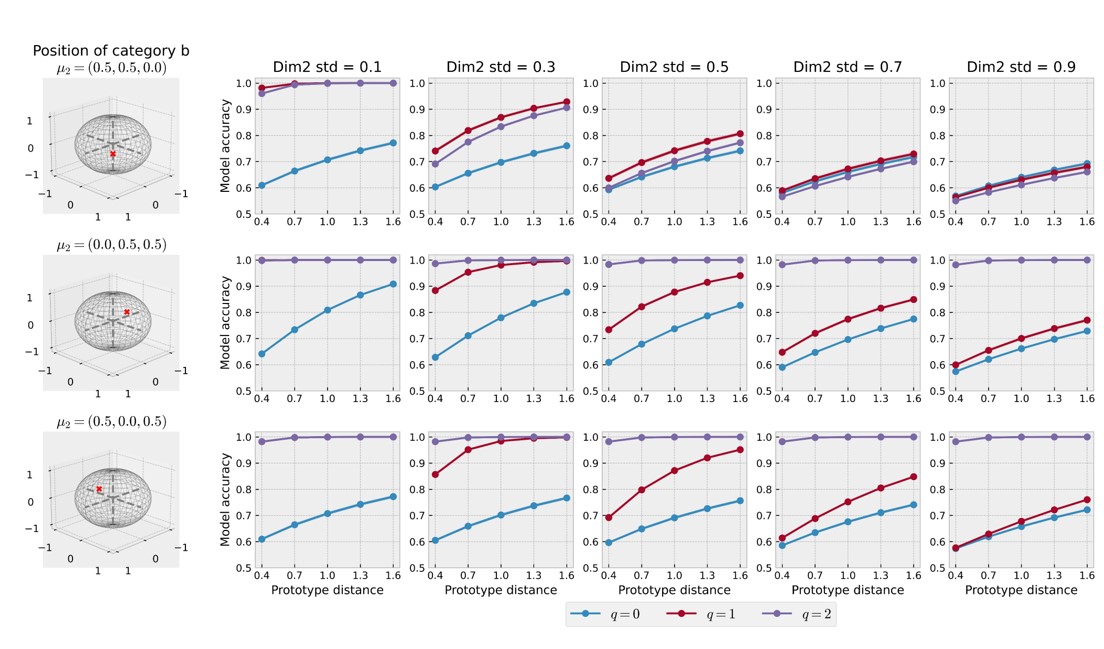}
        \caption{Two informative dimensions}
    \end{subfigure}
    
    \begin{subfigure}{0.9\textwidth}
        \centering
        \includegraphics[width=\linewidth]{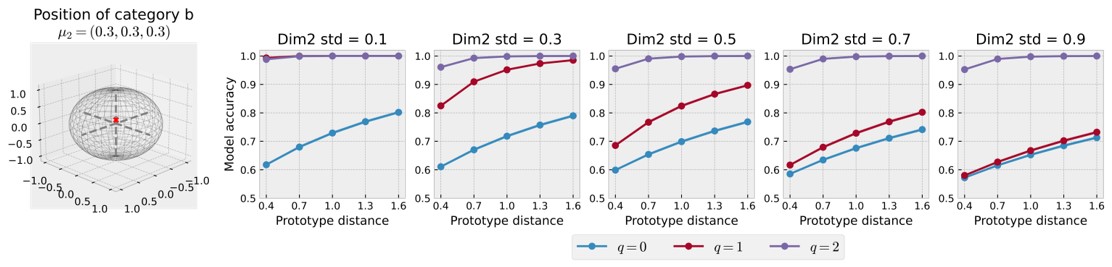}
        \caption{Three dimensions equally informative}
    \end{subfigure}
    \caption{Simulation results. The left column marks the location of category $b$ relative to category $a$. Each figure on the right compares expected accuracy of models with different dimensionality $q$, as distances increases. Figures on in each row demonstrates how increasing the noisiness of the second dimension affects categorization.}
    \label{sim-dimension_reduced_representation}
\end{figure}

Here we present the results in the first simulation study in \cref{sim-dimension_reduced_representation}, demonstrating when will dimension-reduced category representations be helpful. Each sub-figure corresponds to a distribution of information on the 3 dimensions. In \cref{sim-dimension_reduced_representation}(a), there is only one informative dimension. Two equally informative dimensions are present in \cref{sim-dimension_reduced_representation} (b). All three dimensions provide the same amount of information in \cref{sim-dimension_reduced_representation} (c). 
For each sub-figure, the first column on the left illustrates the relative position of the two categories for the plots in the same row. From left to right, the variance on the second dimension is increased, changing the distribution of within-category variation. In each plot, x-axis represents the distance between category prototypes (mean parameter). Y-axis presents the accuracy of the three models, with dimensionality of category representation $q=0,1,2$. $q=0$ implies adoption of Euclidean distance, $q=1$ leads to a 1-dimensional PPCA representation, and $q=2$ is equivalent to a full-rank category representation. 

\subsection{Details of simulation study: hierarchical learning of generalization biases}

We set the concentration parameters in the CRP $\gamma^*$ and $\gamma$ to 1, reflecting moderate preference for new components. We also set vague priors $\Gamma(1,1)$ on parameters $\alpha_\mu$, $\alpha_v$, as well as the precision $\tau=1/\sigma^2$. We choose variational inference (VI) for posterior inference, a common alternative to the Monte Carlo Markov Chain (MCMC) techniques \citep{neal2000markov,sanborn2010rational}. VI approximates intractable posteriors with a family of tractable distributions with optimization \citep{blei2017variational}. It enjoys better scalability compared with MCMC. In recent years, general-purpose VI algorithms have been proposed \citep{ranganath2014black,kucukelbir2017automatic}.
We adopted pyro \citep{bingham2019pyro}, an expressive, scalable and flexible Probabilistic Program Languages, to implement the models. During inference, we perform greedy search for initialization, and train each model with stochastic VI for 3000 epochs.

\subsection{Details of experiment: categorization of natural images \cref{sec:DeepPPCA}}

\paragraph{Discussion of results} When there is no within-category PC, mPPCA reduces to prototype model (with scaling). It can achive high accuracy since the feature map has make the categories relatively easy to separate. After the PC is introduced, the accuracy drops slightly, but the other two metrics enjoy a significant increase. It suggests that rank-1 mPPCA provides a better characterization of graded human generalization patterns. 

Another phenomenon is that mPPCA with a high-dimensional representation has degenerate performance. This is caused by extremely small variation on the last principal direction. Although convolutional network embeddings may not fully capture the psychological space, incidental regularity of dimensions challenges all distance-based models \citep{pettine2023human}. mPPCA with dimension-reduced category representation can perform better at incidental regularity by giving equal weights to noise dimensions.

\subsection{Details of behavioral experiment \cref{sec:exp-fs_generalization}}

Here we present some detailed results that are not included in the main body due to space constraints. 

\subsubsection{Model setup} To provide predictions of human categorization, our models experience the same set of data. Given both stimuli ${x_n}$ and labels $c_n$, the higher-level mixture is disentangled from the lower level. We obtain MLE of the lower-level mixture, then use variational inference for the high-level mixture, i.e. the global PCs, according to \cref{eq:HIM_PPCA_posterior2}. Because of the task context, participants treat the mentioned categories with equal expectation. We set the base rate term in \cref{eq-CategoryAssign} to equal values, which leads to better prediction for all models. 
In the new category experiment, all stimuli in the train and test phase of session 1 is used to get a more reliable estimate of global PCs. This will not be necessary for the subcategory experiment, since only the local PC is needed. 

We compare our model with the exemplar model \citep{nosofsky1986attention}, prototype model \citep{reed1972pattern}, with and without attention mechanism, and the rational model \citep{anderson1991adaptive} with necessary modifications. All models provide predictions without access to human choice. The attention mechanism scales the original space with a set of dimensional weights, optimized based on cluster variations.  
Prototype model with attention mechanism generates generalization pattern similar to that of hierarchical models \citep{salakhutdinov2012one,sanborn2021refresh}, since the categories are dimension-aligned in the experiment. 

We now describe the necessary modifications on the rational model. Rational model represents a category as a infinite mixture of clusters. For subcategory prediction, we assume rational model treats subcategory as one of its clusters. The model first use the instruction to identify the subcategory as one of the clusters. Then for each new sample $y$, we estimate the probability of it belonging to each cluster
\[P(y\in \text{Subcategory}\vert x_\text{sub})=\sum_k P(k\vert x_\text{sub})P(y\vert k),\]
where $k$ indicates clusters within the category.

Meanwhile, we cannot estimate the new category's covariance with one sample. Using a prior on cluster covariance is not fair since category and cluster belong to different levels. As a result, we consider similarity by calculating the sum of similarity to clusters of that category. This is similar to the varying abstraction model \citep{vanpaemel2005varying}.

\subsubsection{Session 1: learning} 
\begin{figure}[htbp]
    \centering
    \begin{subfigure}{0.3\textwidth}
        \centering
        \includegraphics[width=\linewidth]{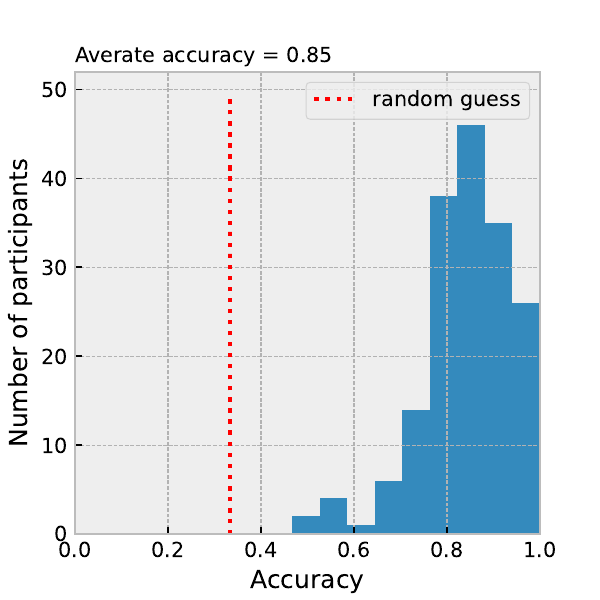}
        \caption{subcategory}
    \end{subfigure}
    \begin{subfigure}{0.3\textwidth}
    \centering
        \includegraphics[width=\linewidth]{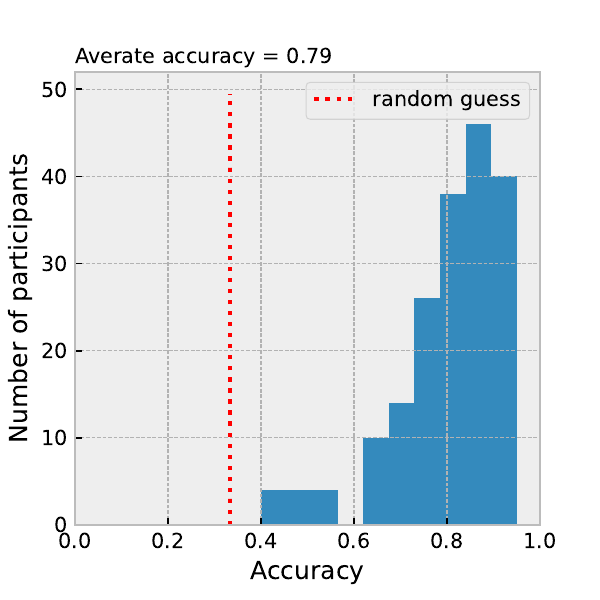}
        \caption{new category}
    \end{subfigure}
    \caption{Subject categorization test accuracy in Session 1. The majority of participants learn the new subcategory or the new category effectively.}
    \label{exp-fs_generalization-session1-test}
\end{figure}

The training and testing phases helps the participants familiarize the stimuli, and learn the category structure in this artificial environment. \cref{exp-fs_generalization-session1-test} shows that the subjects have indeed learned the categories, with accuracy significantly surpassing random guess.

\subsubsection{Session 3: generalization} Test on generalization of categories is the main part of the experiment. Here we present the model predictions on the test stimuli, given the same training experience as human participants. 

\cref{exp-fs_generalization-ProbPred} plots the predicted probability of assigning test stimuli to the subcategory (\cref{exp-fs_generalization-ProbPred_sub}) and new category (\cref{exp-fs_generalization-ProbPred_new}) against human assignment probability. For the \textbf{subcategory experiment}, exemplar models systematically underestimate the probability of the subcategory. Notice that even for the quantitatively best-performing PPCA, there is some under-estimate of assignment probability, especially on those stimuli with human assignment probability around 0.5. We consider this may be an effect of task context. Given specific instruction in \textbf{Session 2} about the existence of a subcategory, participants may naturally tend to choose the subcategory, when they are actually uncertain about the category membership. Prototype models exhibit complex nonlinear patterns. They cannot capture human generalization with flexible switching between contexts. For the \textbf{new category} experiment, the exemplar models again underestimates the probability of the new category. Prototype models provide similar predictions, but generally deviates more from the "ground truth". 

\begin{figure}[htbp]
    \centering
    \includegraphics[width=\linewidth]{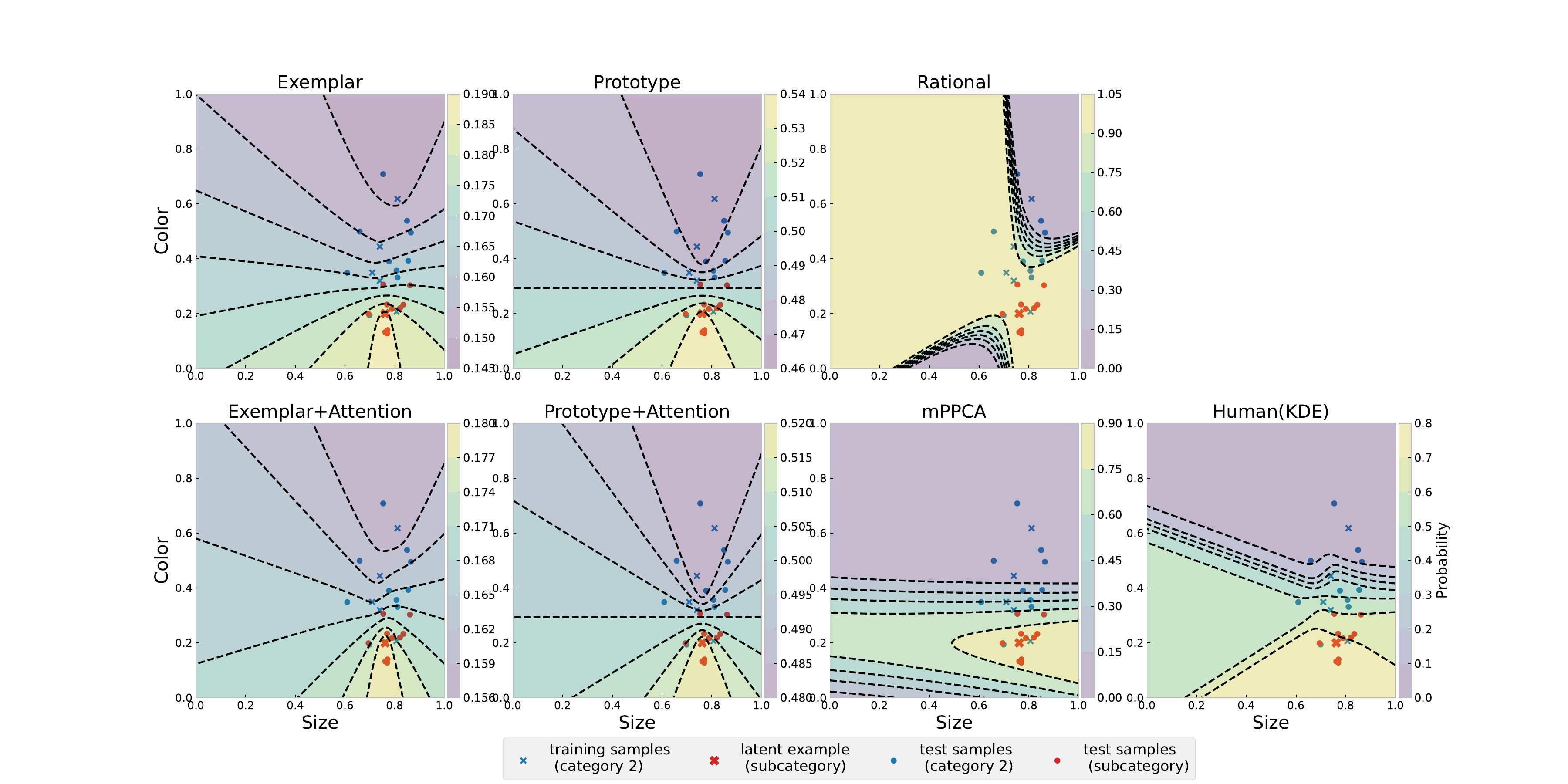}
    \caption{Prediction of generalization pattern in one-shot generalization of subcategory. Dashed lines represent equal generalization probability, dots are the training and generalization test exemplars.}
    \label{exp-fs_generalization-sub_pred}
\end{figure}

\begin{figure}[htbp]
    \centering
    \includegraphics[width=\linewidth]{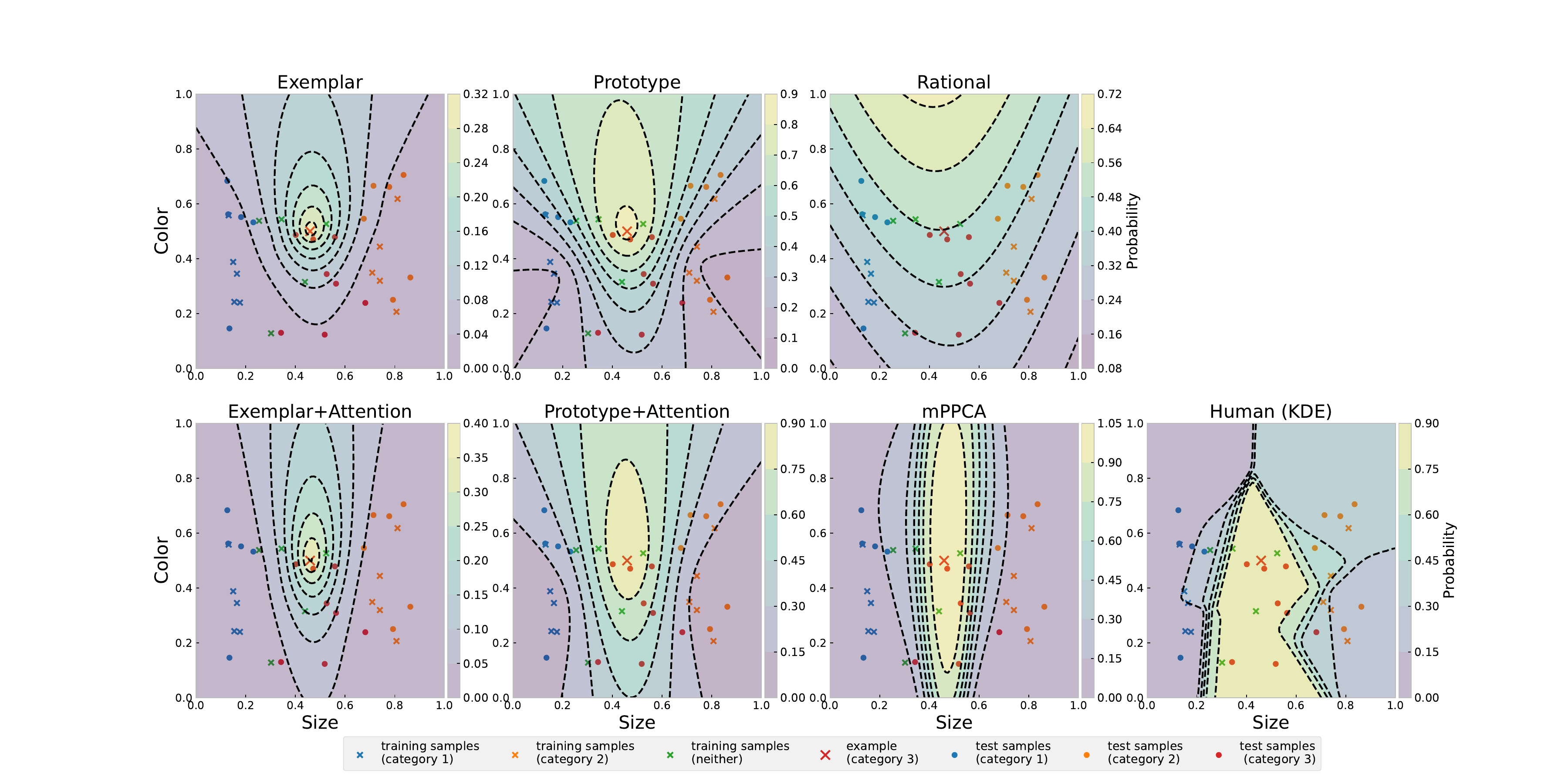}
    \caption{Prediction of generalization pattern in one-shot generalization of new category. Dashed lines represent equal generalization probability, dots are the training/generalization test exemplars.}
    \label{exp-fs_generalization-new_pred}
\end{figure}

\cref{exp-fs_generalization-sub_pred} and \cref{exp-fs_generalization-new_pred} provides the generalization gradients of the subcategory and new category, respectively. In the subcategory experiment, rational model fails to identify the subcategory. This is because learned clusters are not aligned with the subcategory. Exemplar and prototype-based models cannot adjust to the category context flexibly. mPPCA matches human behavior quite well. In the new category setting, mPPCA is similar to prototype with attention. However, we point out that a fixed set of attention cannot account for human categorization. Therefore, mPPCA stands out in explaining human categorization patterns in our experiment.

\begin{figure}[htbp]
    \centering
    \includegraphics[width=0.75\linewidth]{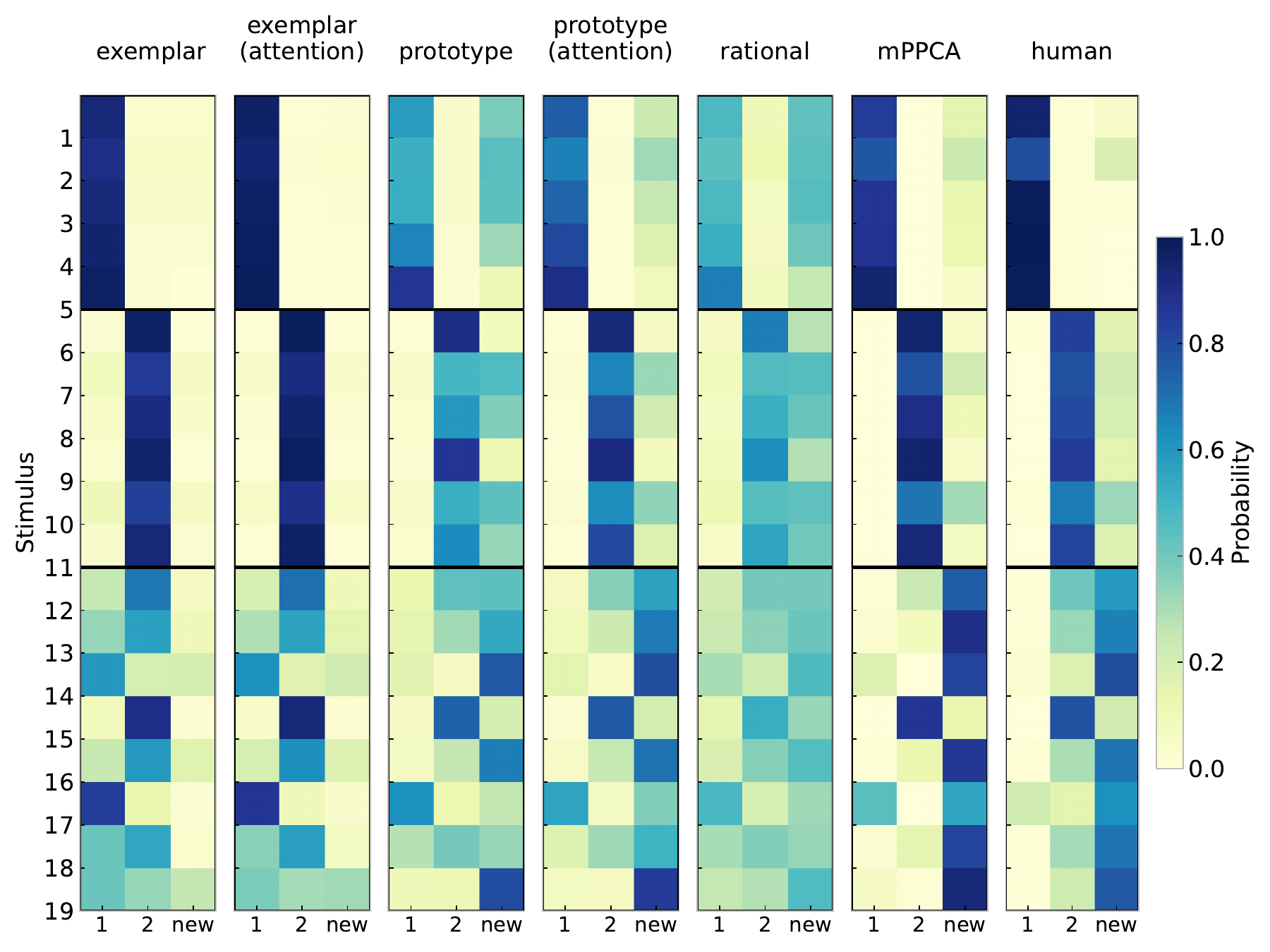}
    \caption{Stimulus-category similarity heatmaps of models in the new category experiment. On the right most is human choice probability.}
    \label{exp-fs_generalization-heatmap}
\end{figure}

For the new category experiment, we also use heatmap in \cref{exp-fs_generalization-heatmap} to illustrate categorization patterns. We can see that mPPCA and prototype model (with attention) provide predictions similar to human categorization probability. Without attention mechanism, prototype model fails to focus on important dimensions for the current task. Exemplar models, on the other hand, underestimates the probability of the new category.

\begin{figure}[htbp]
\centering
    \begin{subfigure}{0.45\textwidth}
        \centering
        \includegraphics[width=\linewidth]{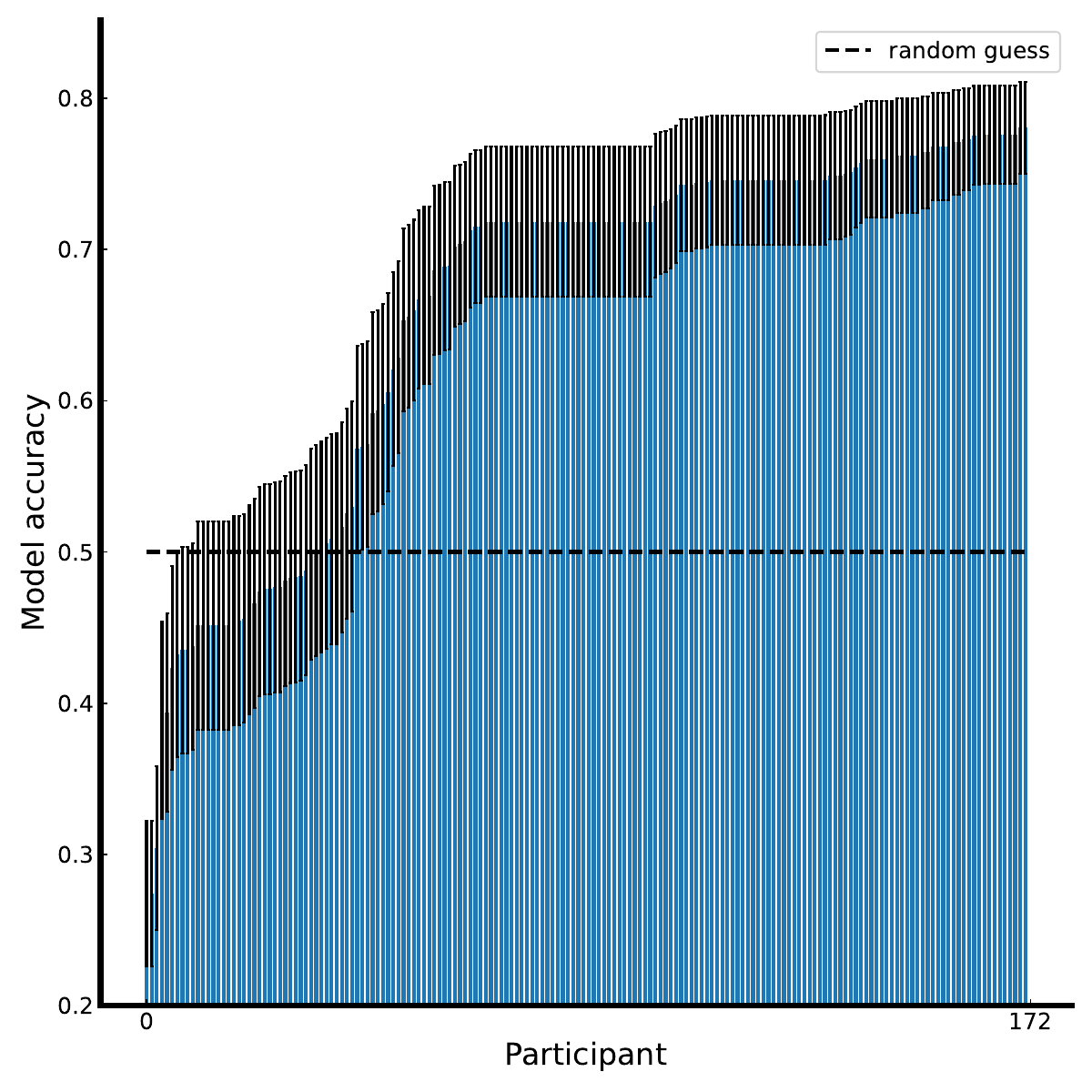}
        \caption{Model accuracy for each individual participant in the subcategory experiment.}\label{exp-fs_generalization-sub_individual_acc}
    \end{subfigure}
    \begin{subfigure}{0.45\textwidth}
        \centering
        \includegraphics[width=\linewidth]{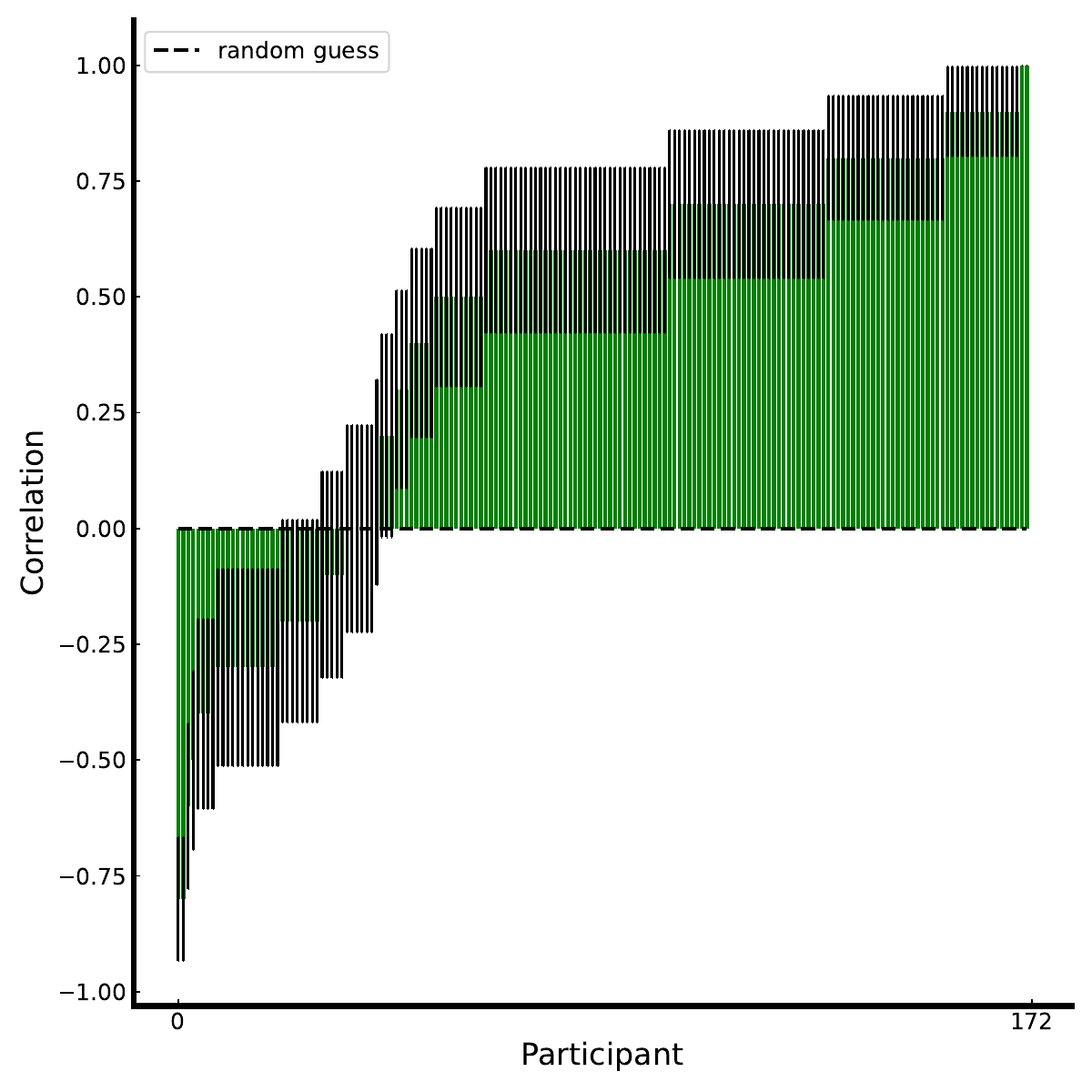}
        \caption{Model correlation to each individual participant in the subcategory experiment.}\label{exp-fs_generalization-sub_individual_cor}
    \end{subfigure}
        \begin{subfigure}{0.6\textwidth}
        \centering
        \includegraphics[width=\linewidth]{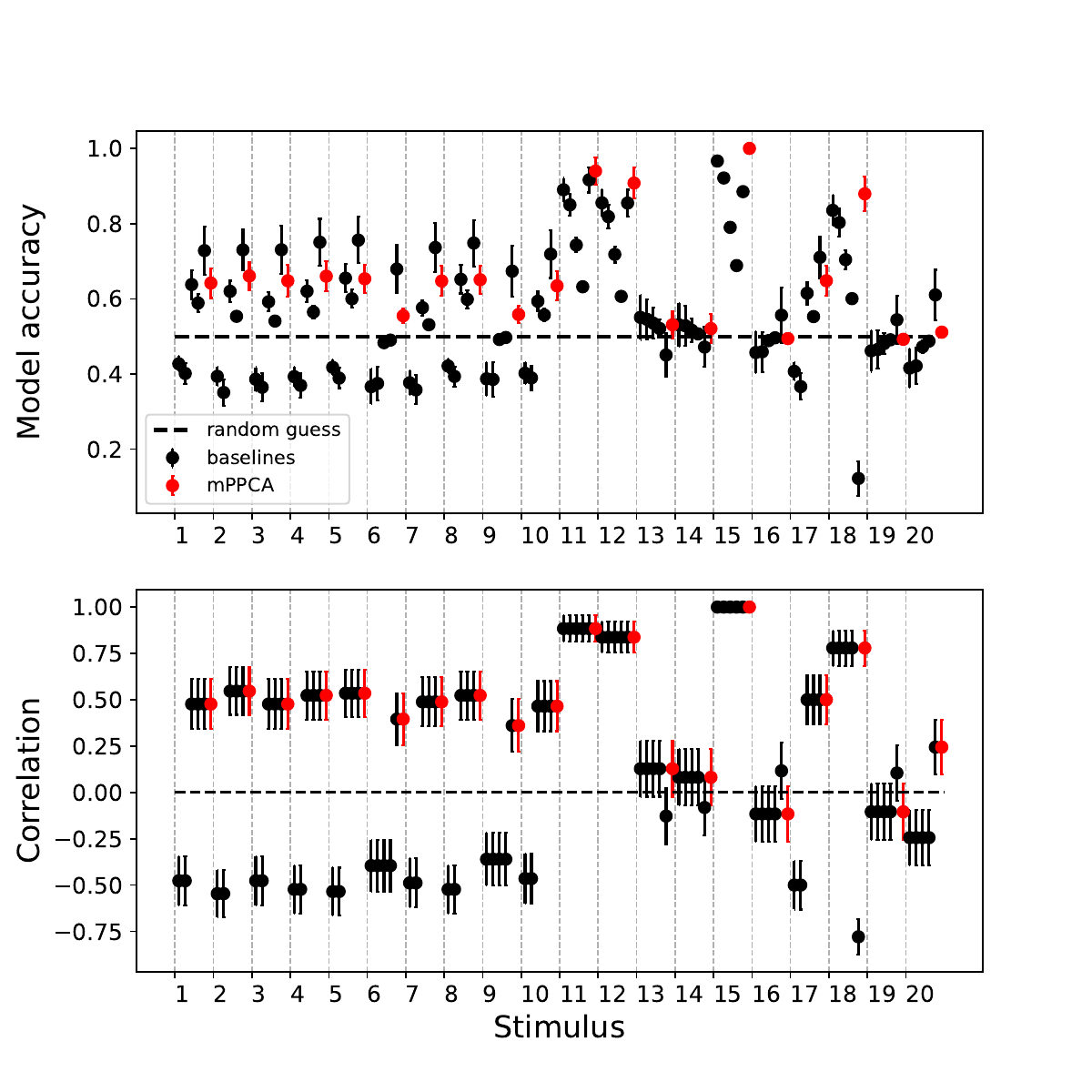}
        \caption{Model prediction performance on each stimulus.}\label{exp-fs_generalization-sub_stimulus}
    \end{subfigure}
    \caption{Participant and stimulus-level analysis of the subcategory experiment.}
    \label{exp-fs_generalization-sub_individual_metric}
\end{figure}

\begin{figure}[htbp]
\centering
    \begin{subfigure}{0.45\textwidth}
        \centering
        \includegraphics[width=\linewidth]{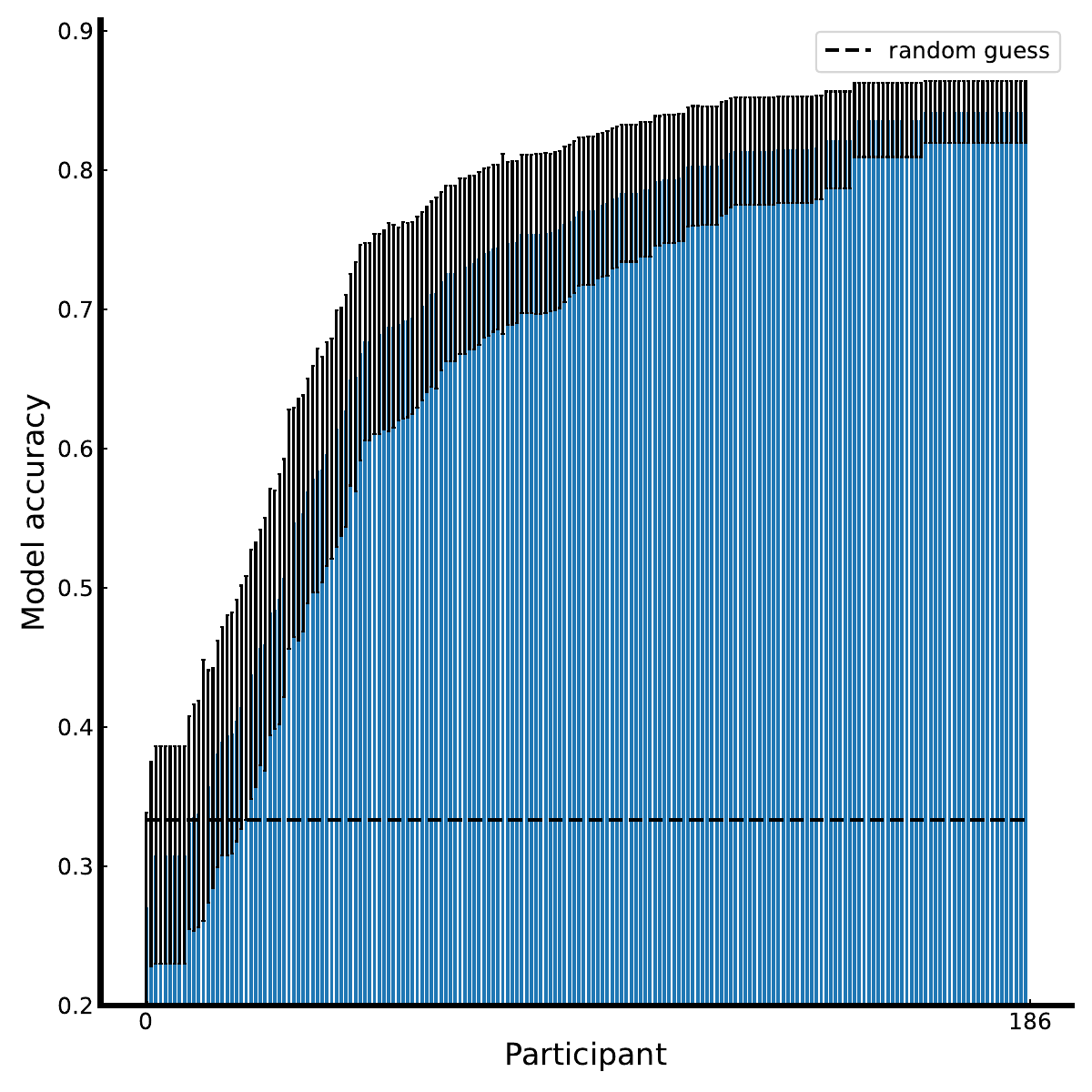}
        \caption{Model accuracy for each individual participant in the subcategory experiment.}\label{exp-fs_generalization-new_individual_acc}
    \end{subfigure}
    \begin{subfigure}{0.45\textwidth}
        \centering
        \includegraphics[width=\linewidth]{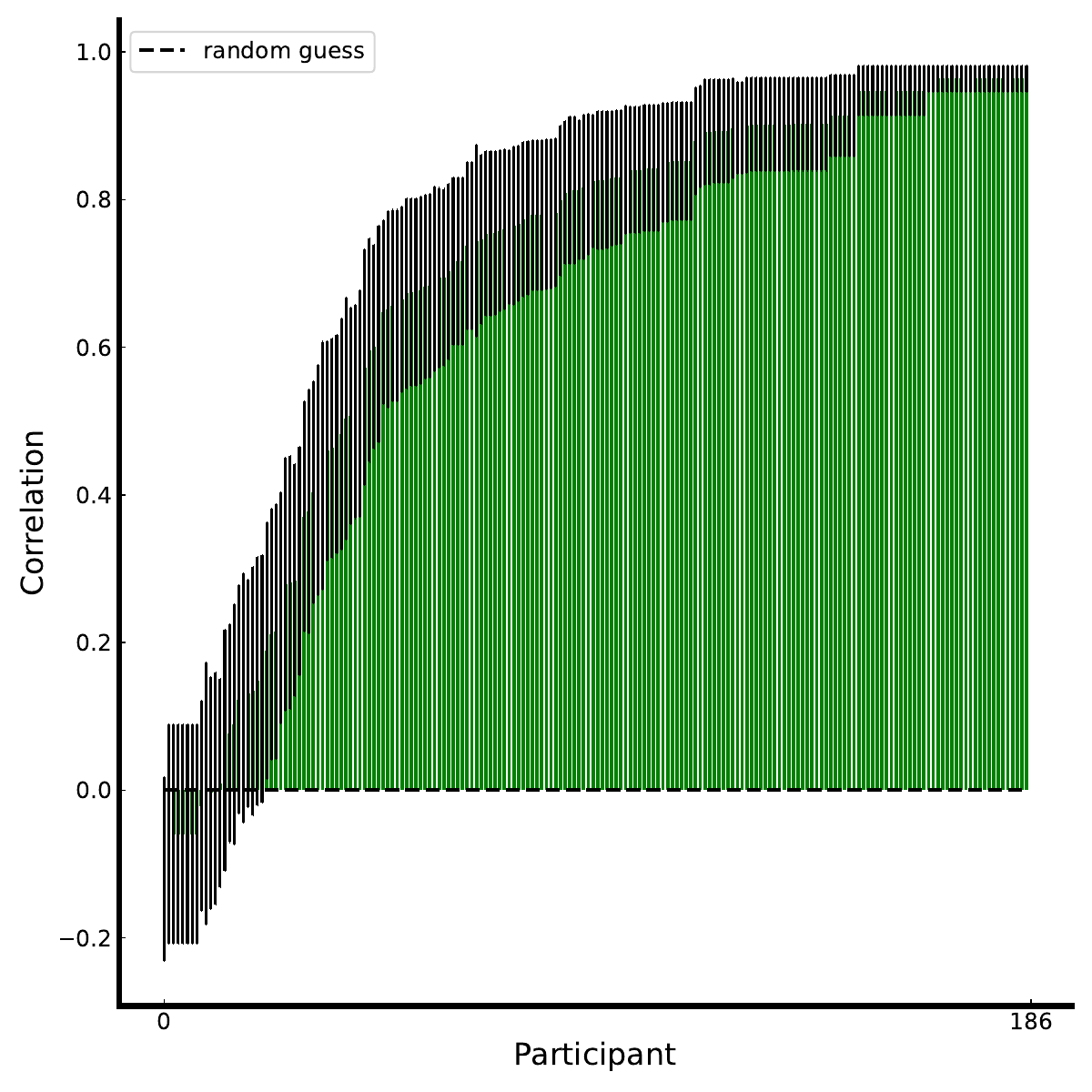}
        \caption{Model correlation to each individual participant in the new category experiment.}\label{exp-fs_generalization-new_individual_cor}
    \end{subfigure}
        \begin{subfigure}{0.6\textwidth}
        \centering
        \includegraphics[width=\linewidth]{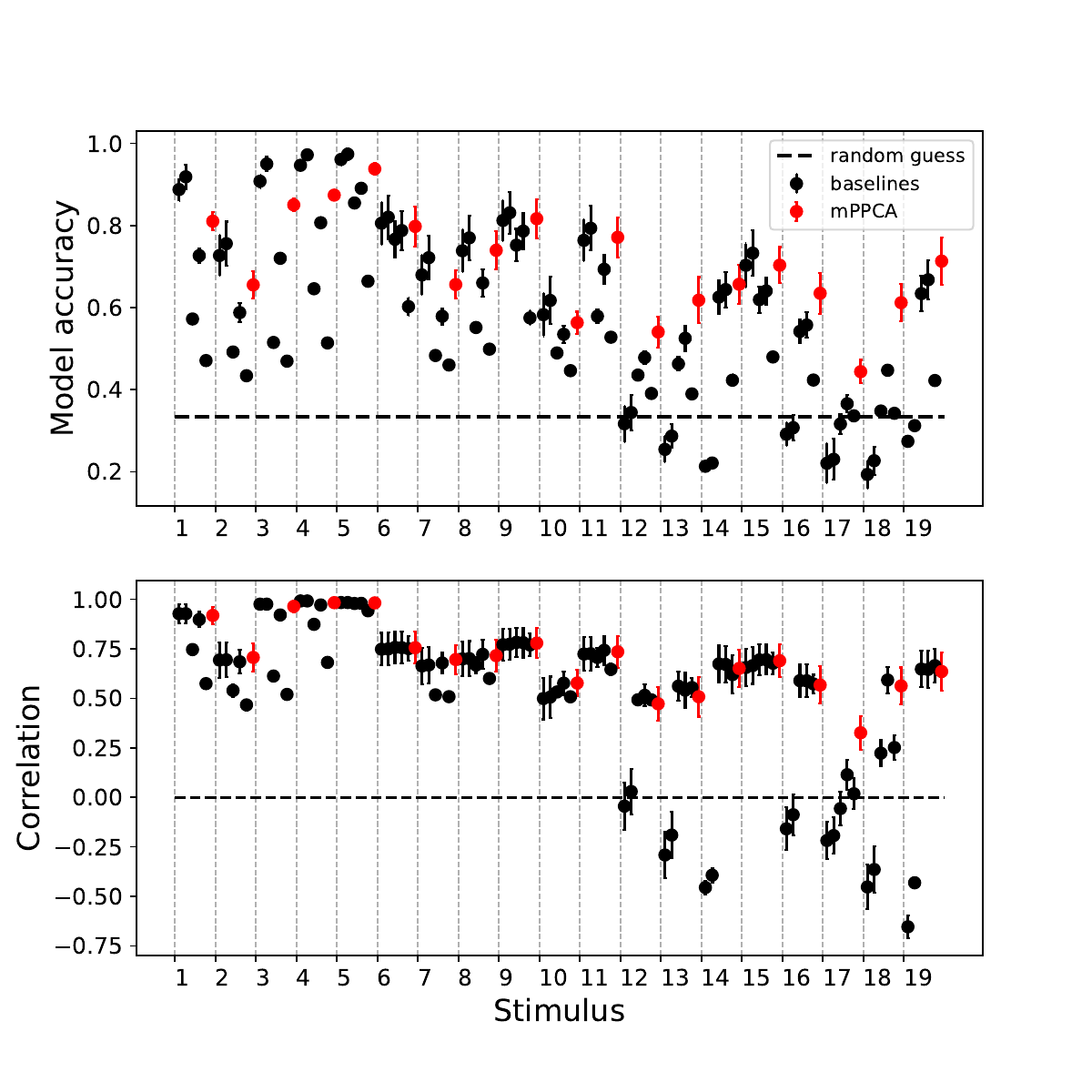}
        \caption{Model prediction performance on each stimulus.}\label{exp-fs_generalization-new_stimulus}
    \end{subfigure}
    \caption{Participant and stimulus-level analysis of the new category experiment}
    \label{exp-fs_generalization-new_individual_metric}
\end{figure}

For more detailed analysis, we show prediction performance for each participant, and each randomly generated stimuli. \cref{exp-fs_generalization-sub_individual_acc} and \cref{exp-fs_generalization-new_individual_acc} illustrate the expected accuracy of mPPCA when predicting human choice probability on the subcategory and new category experiment, respectively. \cref{exp-fs_generalization-sub_individual_cor} and \cref{exp-fs_generalization-new_individual_cor} show the correlation with human categorization on the subcategory and new category experiment, respectively. mPPCA provides a good estimation in these two experiments.

\cref{exp-fs_generalization-sub_stimulus} and \cref{exp-fs_generalization-new_stimulus} show the expected accuracy on for each randomly generated stimulus in the subcategory and new category experiments. In the subcategory experiment, mPPCA performs at least comparably with other models, and is significantly better on some of them. In other words, mPPCA dominates the baseline models, both in terms of accuracy and correlation.
In the new category experiment, mPPCA is outperformed by the exemplar models on simuli from category 1 and category 2. This is caused by the bias of exemplar models towards these categories, which have more training samples. In general, mPPCA provides a better account of human categorization pattern.

\newpage

\section*{Impact statements}\label{sec:impact_statement}
This study share many of the potential societal impacts as other computational cognitive science research. This study focuses on the human behavior of categorization. The major goal of this study is to better the understanding of human mind using computational models. It is necessary to guard against intentional manipulation of humans with the insight provided by cognitive science studies. The gravity of this issue may not be obvious for the current study, but because categorization is a fundamental cognitive activity, we believe it is critical to be cautious about the abuse of scientific discoveries.

During our behavior experiment, human participants were recruited online. The experiment it self has minimal risk. We followed existing protocols and went through the informed consent procedure. The participants are aware of the procedure and can withdraw at anytime. They receive fair bonuses for participation. They allow the data to be used for the present study. Private information is not used or disclosed. 

\newpage

\end{document}